\documentclass[lettersize,journal]{IEEEtran}

\usepackage{graphicx}
\usepackage{amsmath}
\usepackage{amssymb}
\usepackage{booktabs}
\usepackage{multirow}
\usepackage{algorithmicx}
\usepackage{algpseudocode}
\usepackage[caption=false,font=normalsize,labelfont=sf,textfont=sf]{subfig}
 \usepackage{hhline}
 \usepackage{colortbl}
 \usepackage{graphics}
\usepackage{mathptmx}
\usepackage{makecell}
\usepackage[T1]{fontenc}
\usepackage{bbding}
\usepackage{wasysym}
\usepackage{amssymb}
\usepackage{footnote}
\usepackage{colortbl}
\usepackage{mathrsfs}
\usepackage{multirow}
\usepackage{footmisc}
\usepackage{tablefootnote}
\usepackage[table,xcdraw]{xcolor}

\usepackage{amssymb}
\usepackage{pifont}

\newcommand{\etal}{\textit{et al}.}
\usepackage{balance}

\usepackage{footnote}
\usepackage{amsmath}
\usepackage{algpseudocode}
\usepackage{algorithm}
\algnewcommand{\IIf}[1]{\State\algorithmicif\ #1\ \algorithmicthen}
\algnewcommand{\EndIIf}{\unskip\ \algorithmicend\ \algorithmicif}
\usepackage{algcompatible}
\usepackage{enumitem}

\usepackage[breaklinks=true,colorlinks,bookmarks=false]{hyperref}
\hypersetup{colorlinks=true,linkcolor=red,citecolor=blue,urlcolor=magenta}

\newcommand{\mypar}[1]{\vspace{0.1cm}\noindent\textbf{#1}.}

\usepackage[capitalize]{cleveref}
\crefname{section}{Sec.}{Secs.}
\Crefname{section}{Section}{Sections}
\Crefname{table}{Table}{Tables}
\crefname{table}{Tab.}{Tabs.}

\begin{document}

\title{Delving Deep into One-Shot Skeleton-based Action Recognition with Diverse Occlusions}

\author{Kunyu Peng$^{1}$, Alina Roitberg$^{1}$, Kailun Yang$^{1}$, Jiaming Zhang$^{1}$, and Rainer Stiefelhagen$^{1}$
\thanks{The research leading to these results was supported by the SmartAge project sponsored by the Carl Zeiss Stiftung (P2019-01-003; 2021-2026).
The authors would like to thank the consortium for the successful cooperation.
\textit{(Corresponding author: Kailun Yang.)}
}
\thanks{$^{1}$Authors are with Institute for Anthropomatics and Robotics, Karlsruhe Institute of Technology, Germany. (E-mail: \{kunyu.peng, alina.roitberg, kailun.yang, jiaming.zhang, rainer.stiefelhagen\}@kit.edu).}
\thanks{Code will be made publicly available at \href{https://github.com/KPeng9510/Trans4SOAR}{Trans4SOAR}.}
}

\maketitle

\begin{abstract}
Occlusions are  universal disruptions constantly present in the real world.
Especially for sparse representations, such as human skeletons, a few occluded points might destroy the geometrical and temporal continuity critically affecting the results. Yet, the research of data-scarce recognition from skeleton sequences, such as one-shot action recognition, does not explicitly consider occlusions despite their everyday pervasiveness.

In this work, we explicitly tackle body occlusions for \textit{S}keleton-based \textit{O}ne-shot \textit{A}ction \textit{R}ecognition (SOAR). We mainly consider two occlusion variants: 1) random occlusions and 2) more realistic occlusions caused by diverse everyday objects, which we generate by projecting the existing IKEA 3D furniture models into the camera coordinate system of the 3D skeletons with different geometric parameters, (\textit{e.g.}, rotation and displacement). We leverage the proposed pipeline to blend out portions of skeleton sequences of the three popular action recognition datasets (NTU-120, NTU-60 and Toyota Smart Home) and formalize the first benchmark for SOAR from partially occluded body poses. This is the first benchmark which considers occlusions for data-scarce action recognition. Another key property of our benchmark are the more realistic occlusions generated by everyday objects, as even in standard recognition from 3D skeletons, only randomly missing joints were considered. We re-evaluate existing state-of-the-art frameworks for SOAR in the light of this new task and further introduce \textit{Trans4SOAR} -- a new transformer-based model which leverages three data streams and mixed attention fusion mechanism to alleviate the adverse effects caused by occlusions. While our experiments demonstrate a clear decline in accuracy with missing skeleton portions, this effect is smaller with \textit{Trans4SOAR}, which outperforms other architectures on all datasets. Although we specifically focus on \textit{occlusions}, \textit{Trans4SOAR} additionally yields state-of-the-art in the \textit{standard} SOAR without occlusion, surpassing the best published approach by $2.85\%$ on NTU-120.
\end{abstract}

\begin{figure}
\includegraphics[width = 0.48\textwidth]{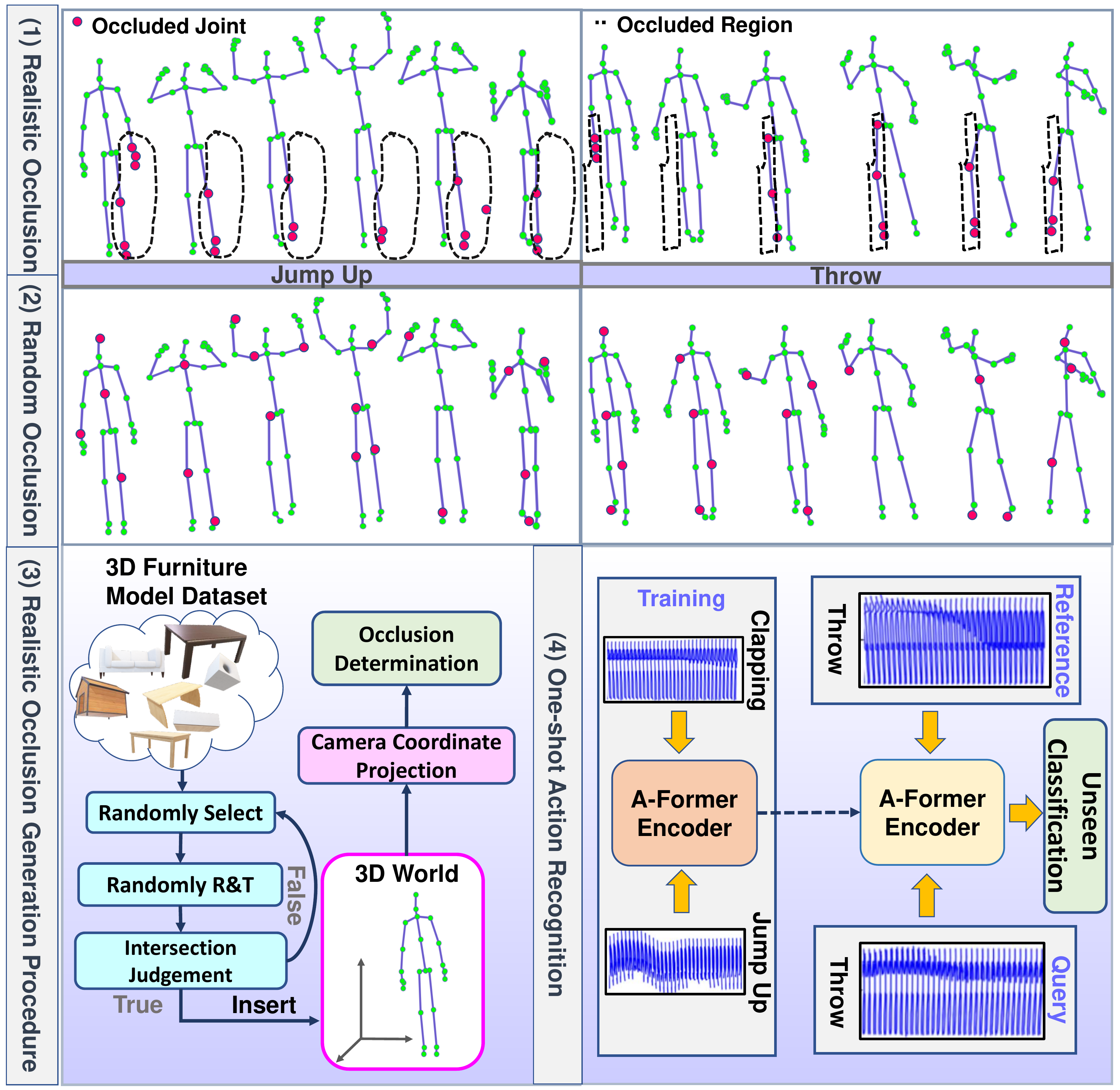}
\caption{An overview of the two proposed and reformulated occlusion scenarios, \textit{i.e.}, REalistic synthesized occlusion (RE) proposed by us as depicted in (1) and RAndom occlusion (RA) reformulated by us depicted in (2). In order to generate realistic occlusion, we randomly choose a 3D furniture model from PIX3D dataset~\cite{sun2018pix3d} and insert it into the world coordinate of the 3D pose using random translation and rotation. The occluded region of the skeletons are determined through camera view point projection, where the whole procedure is demonstrated in (3). In this work we investigate the influence brought by these two dominant occlusions for SOAR, as (4).}

\label{fig:Fig1}
\end{figure} 
\section{Introduction}

\IEEEPARstart{B}{enefiting} from the rapid progress of deep learning, conventional architectures for skeleton-based action recognition achieved remarkable results on a variety of benchmarks for body-pose based classification, \textit{e.g.}, NTU-120~\cite{liu2019ntu} and Toyota Smart Home~\cite{Das_2019_ICCV}.
However, most of the previous works regard relatively clean datasets as the default starting point for training models~\cite{liu2019ntu,zhang2021stst,chen2021channel,yan2018spatial}.
Such datasets often explicitly ensure good body visibility through a suitable camera placement, but this assumption is rather naive, as in real-life the input is often disrupted through \textit{occlusions}.
Skeleton-based action recognition algorithms operate on the coordinates of the 3D body joints and have attracted a great amount of attention~\cite{zhang2021stst,chen2021channel,yan2018spatial,bai2021gcst,cheng2021motion,mazzia2021action,plizzari2021skeleton,plizzari2021spatial,song2022constructing} due to the increasing precision of depth sensors and their privacy-preserving characteristics, but occlusions are especially damaging for such sparse representations, where a few missing joints have a substantial impact on the geometric and temporal continuity.

Learning new concepts with only few labelled examples is often posed in the form of one- or few-shot recognition problem~\cite{cao2021fewshot,gao2020pairwise,hong2021video_pose_distillation,patravali2021metauvfs,perrett2021temporal} and still remains one of the key problems in human action recognition.
Especially if only few training examples are available, occlusions are critical since no diverse data is available for new categories and the quality of the few provided samples dominate the final results.

In this paper, we are interested in categorizing sequences of unseen 3D body poses from only one reference sample, where portions of these sequences are missing due to occlusions. 
Since no past work on one-shot action recognition from skeleton data explicitly considers occlusions, we first introduce a new benchmark by blending out skeleton parts in three established action recognition datasets.  
Our idea is to use a library~\cite{sun2018pix3d} of 3D objects to generate REalistic occlusions (RE), which we project onto the original data with different geometric parameters, such as rotation and displacement.
We believe, that by projecting everyday objects (as shown in Figure~\ref{fig:Fig1}), the occlusions are far more realistic than random dropping of data points, which has been considered in standard~\cite{song2019richly,song2020richly}, (\textit{i.e.}, not data scarce)  recognition from 3D skeletons in the past while preserving the dimension of skeleton joints compared with~\cite{wu2020osd}.
Still, we also consider random blending of body joints while considering both spatial and temporal information as a second occlusion variant, \textit{i.e.}, RAndom occlusion (RA), in our benchmark. 

We also introduce \textit{Trans4SOAR} -- a new transformer-based model which comprises three data streams and mixed fusion to overcome the challenges caused by occlusions. 
Until now, Skeleton-based One-shot Action Recognition (SOAR) has been mostly addressed with Convolutional Neural Networks (CNNs) coupled with metric learning~\cite{memmesheimer2020skeleton_dml,memmesheimer2021sl,zou2018hierarchical} or meta learning~\cite{zou2020adaptation}.
While few recent works considered \textit{transformers} networks in conventional \textit{video}-based human activity classification~\cite{arnab2021vivit}, their potential as signal encoders of \textit{body movement}, their transfer capabilities to \textit{new data-scarce activities classes} and their ability to deal with body pose occlusions have been overlooked. 
For example, Skeleton-DML and SL-DML, the state-of-the-art approaches for the SOAR task~\cite{memmesheimer2020skeleton_dml,memmesheimer2021sl}, both leverage a CNN-based encoder for signal-level skeleton representation and the deep metric learning paradigm.
Our experimental strategy is therefore to first re-evaluate Skeleton-DML~\cite{memmesheimer2020skeleton_dml} and SL-DML~\cite{memmesheimer2021sl} as the current state-of-the-art approaches in the light of our new occlusion-centered task. 
Then, we revisit image-like modelling of skeleton dynamics with the rapidly emerging visual transformers within our \textit{Trans4SOAR} model.
Apart from being the first visual transformer-based architecture for encoding skeleton signals as image-like representations targeting at the SOAR task, we alleviate the adverse effects caused by occlusions through two novel design choices. First, we leverage complementary types of information (velocities, bones and joints) and propose the Mixed Attention Fusion Mechanism (MAFM) which learns how to link the information from diverse streams at the patch embedding level while considering the spatial and temporal neighborhood information.
Secondly, we leverage the Latent Space Consistency (LSC) loss encouraging the model to output similar results with an additional auxiliary branch, if the embedding in the middle layer of the auxiliary branch has been altered by category agnostic prototypes, which suits naturally to the use-case of disturbances through occlusions.

This paper explicitly explores occlusions for SOAR and makes the following contributions:
\begin{itemize}
    \item We for the first time tackle occlusions for \textbf{S}keleton-based \textbf{O}ne-Shot \textbf{A}ction \textbf{R}ecognition (SOAR) and build a benchmark for this task by augmenting three established datasets for action recognition through our occlusion pipeline. Our pipeline considers different occlusion settings, where RAndom occlusion (RA) and REalistically synthesised occlusion (RE) based on everyday objects are the most important ones. We view the latter case as a more practical scenario closer to real-life applications and achieve it by using a library of 3D objects obtained from the IKEA 3D furniture dataset~\cite{sun2018pix3d}, which are inserted into the world coordinate system of the body poses with different geometric parameters.

    \item We introduce \textsc{Trans4SOAR} -- an new three-stream transformer-based model specifically aimed at overcoming data occlusions by 1) leveraging diverse types of input (velocities, bones and joints) and enabling information exchange at the patch embedding level via the Mixed Attention Fusion Mechanism (MAFM) and 2) extensively augmenting the intermediate transformer representations through iteratively estimated category-specific prototypes and the Latent Space Consistency (LSC) loss.
     
    \item  We conduct in-depth experiments in the SOAR task, evaluating two existing state-of-the art frameworks~\cite{memmesheimer2021sl, memmesheimer2020skeleton_dml} as well as our \textsc{Trans4SOAR} network and its individual building blocks under four different types of occlusions. Unsurprisingly, introducing occlusions adversely impacts the outcome, marking an important future work direction. Our \textsc{Trans4SOAR} model yields state-of-the-art on all three datasets under the presence of occlusions.
    
    \item As a side-observation, we discover that \textsc{Trans4SOAR} also outperforms state-of-the-art in the standard SOAR, (\textit{i.e.}, the not occluded SOAR task), surpassing the best previously published model on the challenging NTU-120 SOAR benchmark by $>2.8\%$.
\end{itemize}

\begin{figure*}
    \centering
    \includegraphics[width = 0.95\textwidth]{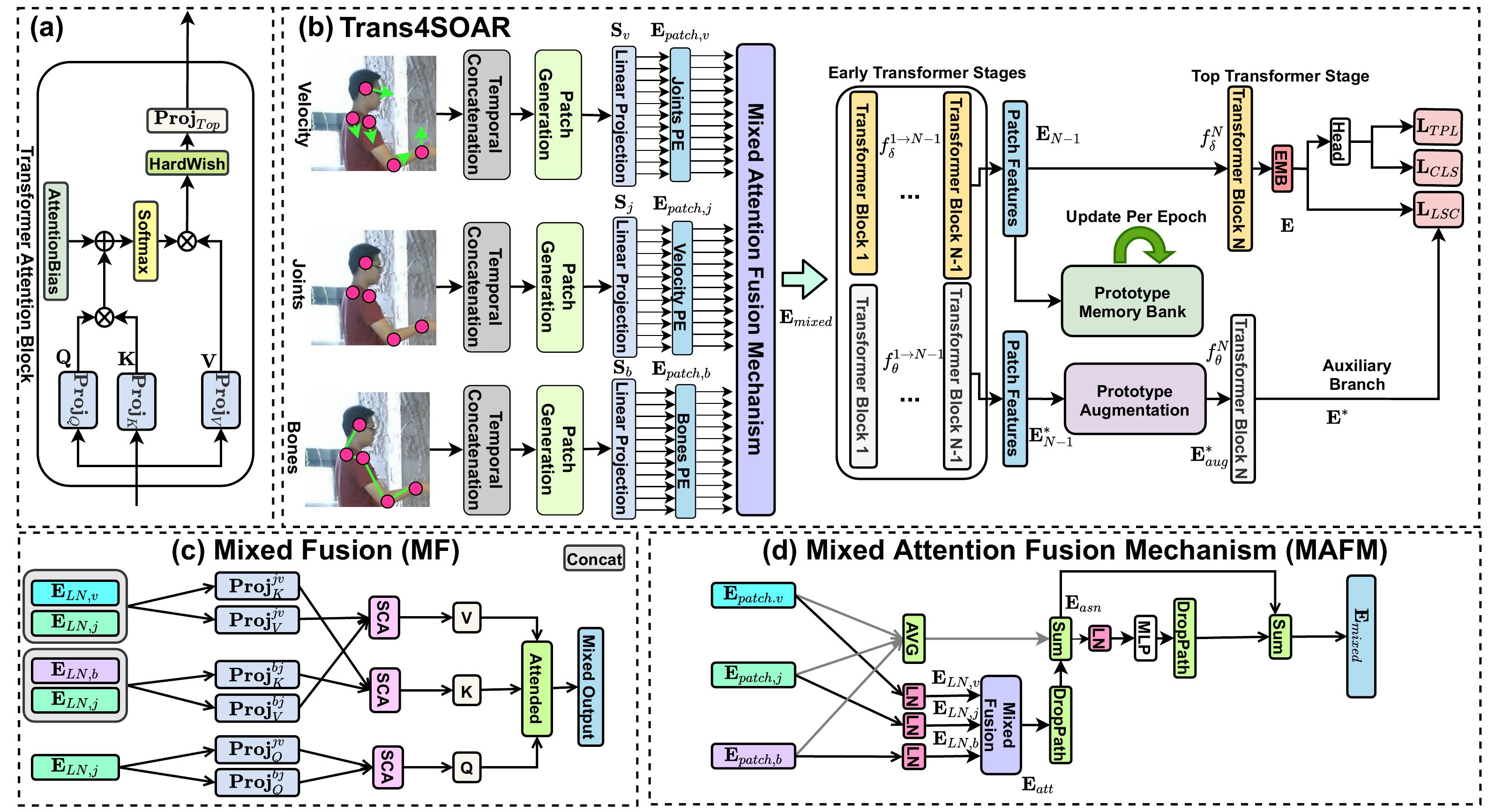}
\caption{An overview of the proposed \textsc{Trans4SOAR} architecture, which is a \textbf{Trans}former for \textbf{S}keleton-based \textbf{O}ne-\textbf{S}hot \textbf{A}ction \textbf{R}ecognition. (a) indicates the transformer block leveraged in \textsc{Trans4SOAR}. This basic transformer attention block is proposed by LeViT~\cite{graham2021levit}, which builds up the transformer block in the later stage of our \textsc{Trans4SOAR} architecture through stacking.
(b) is the overview of the \textsc{Trans4SOAR} training pipeline. First, the skeleton signals are encoded in three kinds of format, \textit{i.e.}, joints, bones, and velocities. Image-like representations are formulated through the concatenation along the temporal axis of the skeleton data, which are further divided into several patches and fed into its corresponding patch embedding net. Then, the Mixed Attention Fusion Mechanism (MAFM) fuses the embeddings from these three different streams by using Mixed Fusion (MF) to achieve cross-stream aggregation on Key, Query, and Value together with the proposed Softmax Concentrated Aggregation (SCA). The Latent Space Consistency (LSC) loss $L_{LSC}$ integrates an prototype augmented auxiliary branch and adopts cosine similarity loss to encourage the embeddings from the main branch $\textbf{E}$ and the embeddings from the auxiliary branch $\textbf{E}^*$ to be more similar. Three losses, \textit{i.e.}, triplet margin loss ($L_{TPL}$), cross entropy loss ($L_ {CLS}$), and LSC loss ($L_{LSC}$), are leveraged for discriminative representation learning. EMB indicates embedding generation layers, which are built based on multi-layer perceptrons (MLP). Head indicates a fully-connected (fc) layer based classification head. PE indicates the patch embedding network. (c) shows the workflow of the Mixed Fusion (MF) and (d) shows the Mixed Attention Fusion Mechanism (MAFM), where \textbf{Proj} indicates the fc-based projection layer, AVG indicates the average operation and LN indicates layer normalization.}
\label{fig:Fig2}
\end{figure*}
\section{Related Work}
\subsection{One-shot Action Recognition}
One-shot recognition, belonged to data-scarce representation learning paradigm, aims at the recognition of unseen category with only one reference samples given as guidance.
Compared with existing works in one-shot image classification, where meta learning-based approaches~\cite{zou2020adaptation,xue2020one,tsutsui2019meta,hu2022pushing,hu2022squeezing,bendou2022easy} dominate the important positions by re-initializing a new task set every epoch following the paradigm regarding learning to learn, Deep Metric Learning (DML) based approaches~\cite{memmesheimer2020skeleton_dml,memmesheimer2021sl,zou2018hierarchical}, which aims at achieving highly discriminative representation and closer representation distance for inter- and intra-category samples in the latent space, are well utilized for Skeleton-based One-shot Action Recognition (SOAR) benchmarked by NTU-120~\cite{liu2019ntu} with pre-defined reference frames. 
One-shot action recognition has been well studied for several down-stream tasks, \textit{e.g.}, semantic segmentation~\cite{zhang2021trans4trans_iccvw} and video classification~\cite{cao2021fewshot,hong2021video_pose_distillation,patravali2021metauvfs,wang2021semantic_guided}, however the research of SOAR are much more sparse and mostly benchmarked on the NTU-120 dataset~\cite{liu2019ntu,memmesheimer2020skeleton_dml,sabater2021one,liu2017skeleton,liu2017global}.
State-of-the-art recognition results are currently reached by the approaches of Memmesheimer~\etal~\cite{memmesheimer2021sl, memmesheimer2020skeleton_dml}, which use a CNN-based encoding of 3D skeletons represented as images and optimizes the framework with deep metric learning using a mixture of cross entropy and triplet margin losses.
In this work, we investigate the transformer architecture and propose a new model named as \textsc{Trans4SOAR} for the SOAR task while considering different occlusion scenarios. We build our optimization paradigm based on SL-DML~\cite{memmesheimer2020skeleton_dml} while further proposing a novel patch embedding level fusion approach considering different skeleton encoding formats.

\subsection{Visual Transformers}
Transformer networks~\cite{vaswani2017attention} are rapidly gaining popularity in computer vision since their operationalization on image patched within the ViT~\cite{dosovitskiy2020vit} and DeiT~\cite{touvron2021deit} architectures. 
Recently, transformer models, known for capturing essential long-range context~\cite{vaswani2017attention}, become increasingly appealing in vision tasks since ViT~\cite{dosovitskiy2020vit} and DeiT~\cite{touvron2021deit} directly utilize a pure transformer on image patches.
A large amount of transformer-based models are thereby put forward regarding this new trend, while some of them target at pursuing better accuracy on image classification task~\cite{chu2021twins,liu2021swin,touvron2021cait}, resource-efficiency~\cite{graham2021levit, zhang2021rest} and the others are designed for more specific tasks, \textit{e.g.}, semantic segmentation~\cite{zhang2021trans4trans_iccvw}.
In action recognition task, the transformer-based models could be arranged into two main groups, which are video-based transformer~\cite{arnab2021vivit,li2021groupformer,zhang2021vidtr,peng2022transdarc} and skeleton-based transformer~\cite{bai2021gcst,mazzia2021action,plizzari2021skeleton,plizzari2021spatial,shi2021star,zhang2021stst} using standard sequential skeleton as input.
We for the first time investigate visual transformer for skeleton data by encoding the skeleton as image-like representation which has the same encoding procedure with SL-DML~\cite{memmesheimer2021sl} while using an additional auxiliary branch and loss to achieve latent space consistency.
Furthermore, in order to achieve high robustness against different occlusion scenarios, a novel feature extraction architecture \textsc{Trans4SOAR} is proposed by integrating a Mixed Attention Fusion Mechanism (MAFM).

\subsection{Skeleton-based Action Recognition with Occlusion}
Most of the skeleton extraction approaches, \textit{e.g.}, AlphaPose~\cite{fang2017rmpe,li2018crowdpose,xiu2018poseflow}, tend to directly give zero output regarding the occluded human body joints.
Thereby some researchers formulated the occluded action recognition task through randomly setting different body regions per frame as zeros to simulate spatial occlusion or setting the randomly selected frame as zeros to simulate temporal occlusion~\cite{song2019richly,song2020richly,li2021partially,ding2020generalized,9732169,8853267} while self-occlusion caused by human body movement is considered in~\cite{8239649}. Notice that, all the aforementioned related works are for skeleton-based action recognition, which is not for SOAR, tackled by our work.
In this work, we jointly consider both spatial occlusion and temporal occlusion at the same time to form random occlusion which is a more reasonable randomly generated occlusion setting since temporal and spatial occlusion should be considered together if the whole skeleton sequence is seen as a sample for SOAR.
Besides, for realistic synthesized occlusion, OSD dataset~\cite{wu2020osd} for the first time projected 3D real world objects into image plane and then generated occluded skeleton in 2D image coordinate. The dimension of raw data is unfortunately not preserved which results in massive information loss by converting 3D data into 2D data.
In order to tackle the occlusion issue in more realistic scenario while preserving dimension of the data, we thereby propose a dimension preserving realistic synthesized occlusion pipeline using an additional 3D model dataset PIX3D~\cite{sun2018pix3d}. Moreover, we for the first time investigate SOAR under diverse occlusions while all the existing works target not-occluded one-shot action recognition.
\section{Benchmark}
\label{sec:benchmark}
To address the lack of related benchmarks, we collect and publicly release the first testbed for SOAR \textit{under presence of occlusions}. 
Our benchmark augments three prominent datasets for SOAR with different occlusions, of which random spatiotemporal occlusions and the realistically synthesised occlusions derived from an existing 3D library of furniture objects being the most important ones.
Next, we give a formal definition of the addressed task (Section \ref{sec:problem}) and 
describe the data obstruction mechanism that we have developed to reach our design goal of realistic everyday occlusions (Section \ref{benchmark:realistic}) as well as the more conventional random occlusion pipeline (Section \ref{benchmark:random}).

\subsection{Problem Definition}
\label{sec:problem}
The task we address is SOAR~\cite{memmesheimer2020skeleton_dml} where a priori knowledge acquired from data-rich action classes is transferred to categorize new data-scarce classes, while certain \textit{regions of the skeleton are not visible}.
Following the standard evaluation protocol for data-scarce action recognition, we build on the one-shot evaluation setting of  NTU-120~\cite{liu2019ntu}, where new categories of the body pose sequences are categorized from a single reference sample.
Formally, $C_{base}$ denotes the set of $|C_{base}|$ data-rich categories available during training through large amount of labelled data $D_{base} = \{(\mathbf{S}_i, l_i)\}_{i=1}^{U}$, $l_i\in C_{base}$ while $U$ indicates the number of samples in $D_{base}$.
Our goal is to distinguish the $|C_{novel}|$ new activity classes $C_{novel}$, for which only $\kappa = 1$ reference training example is available for each class.
These data-scarce examples are referred to as support set $D_{supp} = \{\mathbf{S}_i\}_{i = 1}^{O}$ while $O$ indicates the number of samples in $D_{supp}$ and $C_{base}\cap C_{novel} = \emptyset$.
The final task is then to assign a category $l_n\in C_{novel}$ to each sample from the test set $D_{test}$ containing examples from the data-scarce categories $C_{novel}$.

Since our idea is to study and address \textit{occlusions} for skeleton-based recognition with little training data, we augment both, the support set $D_{supp}$ and the test set $D_{test}$ with our pipeline described in the upcoming sections. 
Note, that in our experiments, we consider both: 1) obstructing the reference examples from the support set and the test set examples and 2) considering occlusions in the test set only, while using complete sequences as our reference samples.

\subsection{Realistic Synthesized Occlusion}
\label{benchmark:realistic}

\begin{figure}[t]
\begin{center}
\includegraphics[width = 0.47\textwidth]{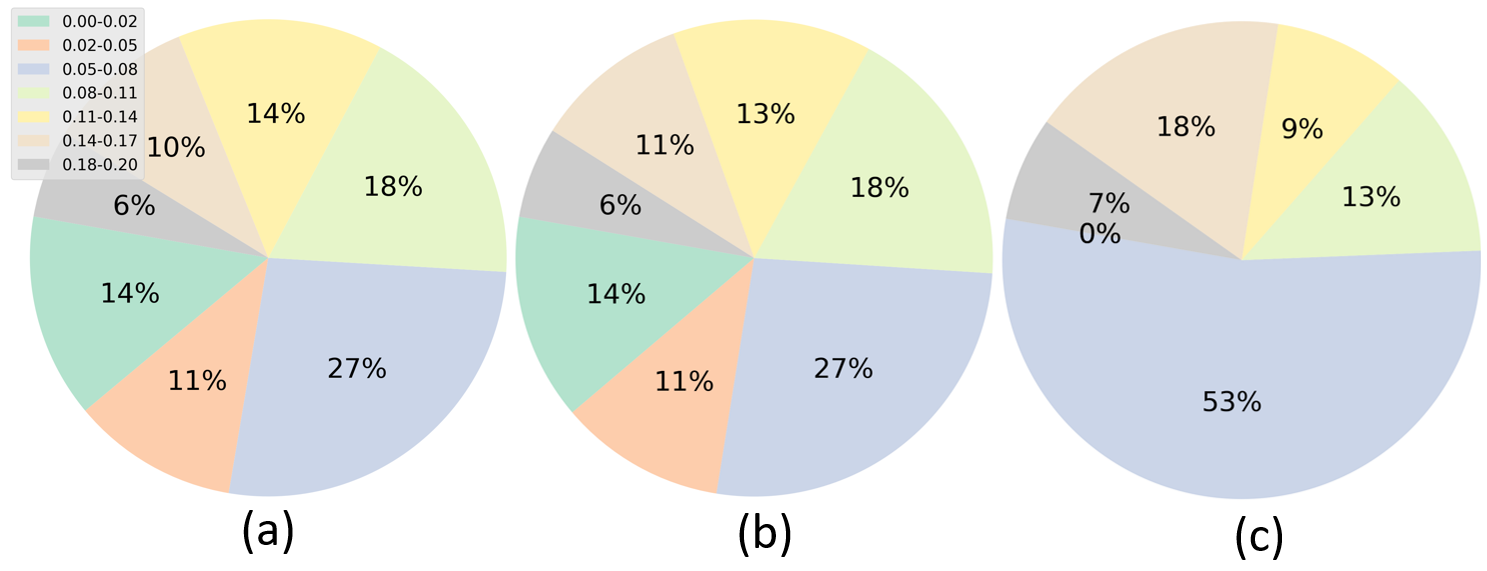}
\end{center}
\caption{An overview of Signal-to-Noise Ratio (SNR) distribution of the realistic synthesized occlusion dataset, where (a), (b) and (c) are for the NTU-120~\cite{liu2019ntu}, the NTU-60~\cite{shahroudy2016ntu} and the Toyota Smart Home~\cite{Das_2019_ICCV} respectively. The legend indicates the corresponding SNR range.
}
\label{fig:statistic_synthesized}
\end{figure}

Past work on skeleton-based activity recognition (without the data scarcity constraint) considered occlusions as random temporal or spatial obstructions only~\cite{song2019richly}, achieved by randomly setting a fixed number of frames or a fixed number of joints to zero respectively.
Such occlusions have highly controllable characteristics but are rather unusual in the real world, where objects are a common cause for obstructions and missing skeleton points exhibit specific geometric consistencies.
To tackle this issue, we build new occluded versions of three public datasets, \textit{i.e.}, NTU-120~\cite{liu2019ntu}, NTU-60~\cite{shahroudy2016ntu} and Toyota Smart Home~\cite{Das_2019_ICCV}, by inserting the 3D IKEA furniture models obtained from the PIX3D dataset~\cite{sun2018pix3d} into the world coordinates system of the human body.
Note, that while NTU-120~\cite{liu2019ntu} and NTU-60~\cite{shahroudy2016ntu} contain 3D skeletons, while Toyota Smart Home~\cite{Das_2019_ICCV} covers 2D data, the process of augmenting the data with the realistic synthesized occlusions is different and will be explained in the following sections.
The statistics of Signal-to-Noise Ratio (SNR) for each dataset with the proposed realistic synthesized occlusion is depicted in Figure~\ref{fig:statistic_synthesized}.

\noindent{\textbf{3D realistic synthesized occlusion dataset generation.}}
The NTU-120~\cite{liu2019ntu} and NTU-60~\cite{shahroudy2016ntu} datasets cover different camera views.
We therefore need to consider the cross-view consistency when obstructing the body poses with the furniture models. 
Unfortunately, the datasets do not provide the calibration data of the individual cameras, which would be the first essential piece of information when addressing this problem.
Luckily, our goal is the skeleton data and each frame containing a skeleton provides sufficient world coordinate information about joints, which can be directly used to calibrate the relative position of the different cameras.
A single skeleton sequence sample with $T$ frames and $J$ joints gives us a set of coordinates with number of $T\cdot J$.
If $N$ samples are provided in our dataset, the known number of the world coordinates thereby rises to $N*T*J$ which is much higher than the rank of projection matrix between two different cameras. 
The calibration matrix $\textbf{F}_{ij}$ between two utilized cameras $i$ and $j$ can be estimated through the following equation:
\begin{equation}
    \mathbf{F}_{ij} = (\mathbf{X}^{T}_{i} \mathbf{X})^{-1}_{i}\mathbf{X}^{T}_{i} \mathbf{X}^{'}_{j}
\end{equation}
where $\mathbf{X}$ denotes the collection of human body joints with homogeneous coordinate format captured simultaneously by two different cameras. 
Thereby, the projection matrix between every two cameras can be obtained as $\textbf{F}_{ij}\in \textbf{F}$. 

A detailed description of the 3D occlusion generation procedure is formalized in Alg.~\ref{algorithm}. 
First, we randomly select 3D object model from the existing IKEA furniture library~\cite{sun2018pix3d} containing 395 different models from 9 categories and augment it via random rotation and translation regarding the vertical axis (bottom to up) and the horizontal plane respectively to simulate life-like occlusions while trying to ensure the bottom points of the skeleton and furniture are in the same vertical height level through vertical translation.
To ensure the cross-view consistency, samples collected simultaneously by different cameras share the rotation and translation augmented furniture model by using the projection matrix from the calibration set $F$.
Next, we need to determine which skeleton body joints are blended out by the current occluded object from the perspective of the camera.
The skeleton body joints $\textbf{x}=[x_1,x_2,x_3]$ (where $x_3$ indicates the axis which is parallel to the focus axis of the camera) and the points of the augmented furniture model $\textbf{z}^{'}$ are first horizontally projected along the focus axis of camera into $\textbf{x}^{\star}$ and $\textbf{z}^{\star}$, as $\textbf{x}^{\star} = [x_1^{\star}/x_3^{\star},x_2^{\star}/x_3^{\star}]$ and $\textbf{z}^{\star} = [z_1^{\star}/z_3^{\star},z_2^{\star}/z_3^{\star}]$.

Then, we build up a two-dimensional convex hull based on the projected points regarding $\textbf{z}^{\star}$ and use the following equation to determine if the human body joints fall into the occlusion convex hull from the camera point of view or not:
\begin{equation}
    IsInHull = IsTrue(A\times(\mathbf{P}^T)\leq Tile(-\mathbf{b}, (1,len(\mathbf{P}))),0), 
\end{equation}
where $A$ denotes the construction equation of the 2D convex hull of $\textbf{z}^{\star}$, $\mathbf{b}$ denotes the last convex hull equation, $\mathbf{P}$ denotes the point that needs to be determined, \textit{Tile} indicates whether all array elements along a given axis are able to be evaluated as True or not and $len(\cdot)$ is the coordinate point dimension.
Thereby the $IsInHull$ is a binary indicator, while $True$ marks the point being inside the convex hull and vice versa. Finally we will get a $Mask$ for each queried skeleton data (see Alg.~\ref{algorithm}), and the occluded 3D points are set to zero.
\begin{algorithm}[t]
    \caption{3D realistic synthesized occlusion generation.}
    \label{algorithm}
    \renewcommand{\thealgorithm}{}
    \begin{algorithmic}[1]
        \small{
            \STATEx \textbf{Input:} $F$ -- the set of projection matrix for each camera pair; $X$ -- the set of skeleton data; $Z$ -- the collection of 3D furniture models from PIX3D dataset; $R$ and $T$ -- random rotation and translation augmentations; $X_{Occ}$ -- a empty set for occluded skeleton data; $[a,~b]$ -- predefined occluded SNR range for the acceptance, where a is the lower limitation and b is the upper limitation.
        }
        \FORALL{$\mathbf{x} \in X$:}
        \STATE \textit{Set $Accept = False$.}
        \WHILE{$Accept != True$}
        \STATE \textit{Set $Found = False$.}
        \STATE \textit{Set $N_{Occ} = 0$.}
        \STATE \Comment{$N_{Occ}$ is the occluded sample number for $X_{d+1}$.}
        \WHILE{$Found != True$}
        \STATE
        \textit{Search $\mathbf{X}_{d}$ collected simultaneously with $\mathbf{x}$ from different views}
        \STATE \textit{Extract the calibration set $F_{d}$ for $X_{d}$.}
        \STATE \textit{Randomly select $\mathbf{z}$, where $\mathbf{z} \in \mathbf{Z}$.} 
        \STATE \textit{Obtain augmented $\mathbf{z}$, i.e., $\mathbf{z}^{'}$, by $\mathbf{z}^{'} = R(T(\mathbf{z}))$.} 
        \STATE \textit{Get $Z_{d}$ by applying $\mathbf{f}_d \in F_{d}$ on $\mathbf{z}$.}
        \STATE \textit{Define $Z_{d+1} = Z_d \cup \{\textbf{z}\}$ and $X_{d+1} = X_d \cup \{\mathbf{x}\}$}
        \IF{\textit{$Z_{d+1}$ has no intersection with $X_{d+1}$ for each corresponding element:}}
            \STATE $Found = True$
        \ENDIF
        \ENDWHILE

        \FOR{$(\mathbf{x}_{d}, \mathbf{z}_{d}) \in zip(X_{d+1},Z_{d+1})$}
        \STATE \textit{Horizontally project $\mathbf{z}_{d}$ and $\mathbf{x}_d$ along focus axis of camera $d$ into 2D plane as $\mathbf{z}^{\star}_{d}$ and $\mathbf{x}^{\star}_d$.}
        \STATE \textit{Build up 2D convex hull $\Phi$ of $\mathbf{z}^{\star}_{d}$.}
        \State \textit{$Mask_{d} = IsInHull(\Phi, \mathbf{x}^{\star}_d)$.}
        \State \textit{Calculate $SNR_{d} =Sum(Mask_{d})/len(Mask_{d})$ for $\mathbf{x}_d$.}
        \State \textit{Occlude $\mathbf{x}_{d}$ by $\mathbf{x}_d[Mask_{d}] = zeros\_like(\mathbf{x}_d[Mask_{d}])$}
        \State \textit{Append $\mathbf{x}_d$ into $X_{d}^{Occ}$}
        \IF{\textit{$SNR_{d}$ in $[a,b]$}}
        \STATE \textit{$N_{Occ} += 1$.}
        \ENDIF
        \ENDFOR
        \IF{$N_{Occ}<T_{Occ}$ or $N_{Rep}<T_{Rep}$}
        \STATE \textit{Set $Accept = False$ and $N_{Rep} += 1$.}
        \ELSE
        \STATE \textit{Set $Accept = True$.}
        \STATE \textit{Del $X_d$ from $X$ and append the $X_{d}^{Occ}$ into $X_{Occ}$.}
        \ENDIF
        \ENDWHILE
        \ENDFOR
    \end{algorithmic}
\end{algorithm}

\noindent{\textbf{2D realistic synthesized occlusion dataset generation.}}
Since the third dataset we leverage, \textit{i.e.}, Toyota Smart Home~\cite{Das_2019_ICCV}, only contains 2D skeletons in the image plane, the aforementioned pipeline for generating 3D realistic synthesized occlusion is modified to suit the 2D use-case.
Instead of directly augmenting the 3D furniture model via rotation and translation, we use a randomly generated projection matrix, which transforms the points from the camera coordinates to the image coordinate system, to project the furniture model onto the image plane. 
Then, similar to the 3D realistic synthesized occlusion generation procedure, the convex hull of the occluded region is built up according to the projected points of the furniture model on the 2D image plane and an occlusion-aware mask is obtained through the $ISInHull$ function. 
Finally, the corresponding 2D skeleton joints within the mask are changed to zeros.

\subsection{Random Occlusion}
\label{benchmark:random}

The second occlusion variant we considered is \textit{random} data obstruction, which is a combination of random temporal and spatial occlusions used in past work on standard, (\textit{i.e.}, without the data scarcity constraint) skeleton-based action recognition~\cite{song2019richly,song2020richly}.
For random temporal occlusions, a fixed number of frames are blended out randomly for each skeleton sequence to simulate full occlusion for certain points in time.
For random spatial occlusions, a fixed number of joints are  randomly set to zero in every frame of the skeleton data stream. This is a very specialized type of occlusions, since the exact number of joints are not visible in all frames.
However, mixing both, random temporal occlusions and random spatial occlusions, is a more reasonable scenario with less predefined controllable conditions and a higher chance to happen in real-life.
With $\textbf{x}\in \mathbb{R}^{T\times J \times B}$ being the skeleton data, we first flatten it into $\mathbb{R}^{(T\times J) \times B}$, after which a set of data points $\gamma \cdot (T\times J)$ are randomly chosen with a predefined SNR ratio $\gamma$ and blended out.
Although we view the mixed spatial and temporal occlusion as a more reasonable option, we also conduct experiments with isolated random spatial and temporal occlusions for consistency.
Overall, our experiments described in the later sections will indicate, that the realistic synthesized occlusions are the most challenging ones.

\section{Methods: Trans4SOAR}
We introduce \textit{Trans4SOAR} -- a three-stream transformer-based model designed to overcome adverse effects of occlusions (an overview is provided in Fig. \ref{fig:Fig2}).
The key ingredients of our model are 1) the
Mixed Attention Fusion Mechanism (MAFM) which learns to aggregate three different types of skeleton information,  (\textit{i.e.,} joints, velocities and bones) at the patch embedding level and 2) an extensive data augmentation technique at the feature-level, where an auxiliary branch is augmented by category agnostic prototypes.
The motivation of the latter component is to encourage better robustness against imperfect data brought by occlusions through an additional consistency cost computed between the obtained body pose embedding and its prototype-augmented version.

Next, we describe the basic components regarding the input encoding, the patch embedding procedure and the basic transformer blocks of \textsc{Trans4SOAR} (Sec.~\ref{sec:components_trans4soar}) and MAFM -- the central building block model responsible for the three-stream fusion at the patch embedding level (Sec.~\ref{sec:MAFM}). 
Finally we introduce our auxiliary Latent Space Consistency (LSC) loss for encouraging invariance to  transformations through consistency constraints and augmentations with previously learned action category prototypes (Sec.~\ref{sec:prototype_loss}). 

\subsection{Illustration of the Base Components}
\label{sec:components_trans4soar}
\noindent{\textbf{Input encoding.}}
We follow the skeleton encoding for body joints proposed by SL-DML~\cite{memmesheimer2021sl} to cast the sequential skeleton data as image-like representations.
Assuming that $\mathbf{s}{\in}\mathbb{R}^{T\times J\times B}$ denotes a sequential skeleton sample, where $T$ indicates the temporal length, $J$ indicates the total number of joints, and $B$ indicates the dimension of the coordinates of the skeleton joints.
The input of our approach is achieved by interpolation, which transfers $s$ from $T{\times}J{\times}B$ to $H{\times}W{\times}B$ to match the image-wise input.
Moreover,  \textsc{Trans4SOAR} is a three-stream architecture, which does not only consider joints (as in \cite{memmesheimer2021sl}) but also bones $\textbf{b}$ and velocities  $\textbf{v}$ defined as $\textbf{v}_t{=}\textbf{s}_{t}{-}\textbf{s}_{t-1}$, denoting the velocity for each joint during human body motion at time stamp $t$, and $\textbf{b}_{i, j} {=} \textbf{s}_i{-}\textbf{s}_j,~for~(i,j)\in\Omega_{bones}$, denoting the vector of each bone of the human skeleton, respectively.
These vectors are subsequently mapped to image-like arrays using the described above procedure.

Finally, after the skeleton format encoding and image-wise transformation, we obtain three individually image-wise inputs including joints, velocities, and bones which have the same shape as $H{\times}W{\times}B$. So at the end of input encoding we have three streams of input, \textit{i.e.}, joints, velocities and bones.

\noindent{\textbf{Patch embedding and transformer blocks.}}
Modern CNNs are excellent at preserving details, while transformers are known for capturing long-range dependencies, making the combination of CNN- and transformer blocks a potential double win.
LeViT~\cite{graham2021levit} opened the door for this combination by using a CNN with four layers for patch embedding, before the stack of transformer blocks.
Standing on the shoulders of giants, we adopt the basic transformer blocks and patch embedding layers proposed by LeViT~\cite{graham2021levit} in \textsc{Trans4SOAR}.
The basic transformer attention block of LeViT~\cite{graham2021levit} is depicted on the left hand side of Figure~\ref{fig:Fig2}.
After the acquisition of the Query $\mathbf{Q}$, Key $\mathbf{K}$ and Value $\mathbf{V}$ through the projection layers $\mathbf{Proj}_{Q}$, $\mathbf{Proj}_{K}$, and $\mathbf{Proj}_{V}$ respectively, the final attended output can be calculated as~ $\mathbf{Proj}_{Top}(HardWish((Softmax(\mathbf{Q}\times\mathbf{K})) + Bias_{att})\times \mathbf{V})$, where each projection layer is composed of an 1x1 conv and a batch normalization layer, and $Bias_{att}$ denotes the attention bias.
First, the leveraged three streams of inputs namely joints, velocities, and bones are separately divided into $N_{patch}{=}(H/P){\times}(W/P)$ patches with a predefined path size $P$ and thereby three input sequences are produced, which are denoted as $S_j$, $S_v$, and $S_b$ for joints, velocities, and bones, respectively.
Then, we build up patch embedding layers through a stack of convolutional neural networks, denoted as $M_{\theta^j}$, $M_{\theta^v}$, and $M_{\theta^b}$ to extract high-dimensional embeddings for the patch sequence of each stream, denoted separately as $\textbf{E}_{patch,j}$, $\textbf{E}_{patch,v}$, and $\textbf{E}_{patch,b}$. We follow the attention bias setting (depicted in Figure~\ref{fig:Fig2} (a)) instead of using position embeddings as introduced in~\cite{graham2021levit}.
The corresponding relationship is depicted in Eq.~(\ref{eq:patch_embedding}):
\begin{equation}\label{eq:patch_embedding}
\centering
\mathbf{E}_{patch,j},~ \mathbf{E}_{patch,v},~ \mathbf{E}_{patch,b} = M_{\theta^j}(\textbf{S}_j),~M_{\theta^v}(\textbf{S}_v),~M_{\theta^b}(\textbf{S}_b).
\end{equation}
The resulted three embedding streams are then fed into the key component of the proposed \textsc{Trans4SOAR} architecture, \textit{i.e.} Mixed Attention Fusion Mechanism (MAFM), for multi-stream fusion, which we now introduce in detail.
\subsection{Multimodal Fusion at the Patch Embedding Level}
\label{sec:MAFM}
\noindent{\textbf{Mixed Fusion (MF).}}
The proposed Mixed Attention Fusion Mechanism (MAMF) builds upon the the Mixed Fusion (MF) strategy.
The main purpose of MF is to transfer the important cues from the two auxiliary streams, \textit{i.e.}, velocities and bones, to the main stream, \textit{i.e.}, joints, to achieve multi-stream fusion of skeleton data on the patch embedding level. We propose to use a mixture of Value and Key for multi-stream fusion. While such a concept regarding the mixture of Key and Value is studied in MixFormer~\cite{cui2022mixformer} for template matching, the design of our proposed MF mechanism is non-trivial. Unlike the mixture in template matching which aims to push the model to focus on similarity cues, our MF scheme is designed to harvest complementary cross-modality dependencies and enable a multi-stream agreement for discriminative embedding learning. Apart from using a concept of mixture of Key and Value, we design a unique three-stream patch-embedding fusion architecture regarding MF and MAFM to suit the discriminative embedding learning the for SOAR.
In the following, we introduce the proposed MF for multi-stream patch embedding fusion in detail.
First, we encode the patch embeddings of the joints $\mathbf{E}_{patch,j}$ through two different linear projection layers, \textit{i.e.}, $\mathbf{Proj}_{Q}^{jv}$ and $\mathbf{Proj}_{Q}^{bj}$, as depicted in Eq.~(\ref{eq:linear_projection}):
\begin{equation}\label{eq:linear_projection}
    \mathbf{Q}_{jv},~\mathbf{Q}_{bj} = \mathbf{Proj}_{Q}^{jv}(\mathbf{E}_{patch,j}),~\mathbf{Proj}_{Q}^{bj}(\mathbf{E}_{patch,j}).
\end{equation}
Then, for Keys and Values of the $jv$ branch and the $bj$ branch, the input embeddings are aggregated together through concatenation, which is indicated as $Cat$.
After that, for each single term, a projection layer is used for encoding.
For example, $\mathbf{Proj}_{V}^{jv}$ is the projection layer for Value of the $jv$ branch.
As a result, $\mathbf{V}_{iv}$, $\mathbf{K}_{jv}$, $\mathbf{V}_{bj}$, and $\mathbf{K}_{bj}$ are yielded after the encoding:
\begin{equation}
    \mathbf{V}_{iv} = \mathbf{Proj}_{V}^{jv}(Cat(\mathbf{E}_{patch,j},~ \mathbf{E}_{patch,v})),
\end{equation}
\begin{equation}
    \mathbf{K}_{jv} = \mathbf{Proj}_{K}^{jv}(Cat(\mathbf{E}_{patch,j},~ \mathbf{E}_{patch,v})),
\end{equation}
\begin{equation}
    \mathbf{V}_{bj} = \mathbf{Proj}_{V}^{bj}(Cat(\mathbf{E}_{patch,j},~ \mathbf{E}_{patch,b})),
\end{equation}
\begin{equation}
    \mathbf{K}_{bj} = \mathbf{Proj}_{K}^{bj}(Cat(\mathbf{E}_{patch,j},~ \mathbf{E}_{patch,b})).
\end{equation}
After the aforementioned procedures, we have obtained Query, Key, and Value for the two branches, separately.
Then the question lies in how to further aggregate these two branches.
We introduce Softmax Concentrated Aggregation (SCA), which is realized through the following equations to achieve aggregation between $\mathbf{V}_{jv}$ and $\mathbf{V}_{bj}$, $\mathbf{K}_{jv}$ and $\mathbf{K}_{bj}$, and $\mathbf{Q}_{jv}$ and $\mathbf{Q}_{bj}$:
\begin{equation}
    \mathbf{V} = (Softmax(\mathbf{V}_{jv})^{T}\mathbf{V}_{bj} + Softmax(\mathbf{V}_{bj})^T\mathbf{V}_{jv})/2,
\end{equation}
\begin{equation}
    \mathbf{K} = (Softmax(\mathbf{K}_{jv})^{T}\mathbf{K}_{bj} + Softmax(\mathbf{K}_{bj})^T\mathbf{K}_{jv})/2,
\end{equation}
\begin{equation}
    \mathbf{Q} = (Softmax(\mathbf{Q}_{jv})^{T}\mathbf{Q}_{bj} + Softmax(\mathbf{Q}_{bj})^T\mathbf{Q}_{jv})/2.
\end{equation}
After the SCA operation, we obtain the aggregated Query $\mathbf{Q}$, Key $\mathbf{K}$, and Value $\mathbf{V}$, which is merged together to formulate the desired attention by $Att = Softmax(\mathbf{Q}\mathbf{K}^T/\sqrt{s_k})\mathbf{V}$, where the scale factor is used to avoid the negative influence brought by the dot product on the variance and $Att$ denotes the calculated attention value.
\setlength{\textfloatsep}{5pt}
\begin{algorithm}[t]
    \caption{An overview of the training pipeline with latent space consistency (LSC) loss.}
    \label{algorithm1}
    \renewcommand{\thealgorithm}{}
  
    \begin{algorithmic}[1]
        \small{
            \STATEx \textbf{Input:}  $\textbf{S}$ -- a batch in $D_{train}$; $\textbf{S}_{p}$ and $\textbf{S}_{n}$ -- positive and negative anchor;  $f_{\delta}^{1 \to N-1}$ and $f_{\theta}^{1 \to N-1}$ --  first N-1 transformer layers of main and auxiliary branches; $f_{\delta}^N$ and $f_{\theta}^N$ -- the N-th (last) transformer layer for main and auxiliary branches; $EMB$ -- Embedding layer; $N_e$ -- maximum training epochs; $N_t$ -- epoch threshold for the stage changing; $\textbf{E}$ and $\textbf{E}^*$ -- embedding for main and auxiliary branches; PMB -- Prototypes memory bank; WarmUpAug and PrototypeAug -- Warm-up stage and prototype-based feature augmentation stage
            
        }
        \FORALL{$epoch \in Range(N_e)$}
        \FORALL{$\textbf{S} \in D_{train}$}
                \IF {\textit{$epoch>N_t$}} 
                \FORALL{$l$ \textit{in} $label_S$}
                \textit{Append($PMB[l]$)} $\to$ \textit{$List_p$}
                \ENDFOR
                \STATE \textit{$\textbf{E}_{P}^* =$ Concat($List_p$)}
                \ENDIF
        \IF {\textit{BaseModel is not None}} $\textbf{S}=BaseModel(\textbf{S})$ \ENDIF
        \STATE $\textbf{E}_{patch} = PatchEmbeddingAndEncoding(\textbf{S})$
        \STATE $\textbf{E}_{N-1} = f_{\delta}^{1 \to N-1}(\textbf{E}_{patch})$, $\textbf{E}_{N-1}^*={f_{\theta}^{1 \to N-1}}(\textbf{E}_{patch})$
        \IF {\textit{$epoch<N_t$}} $\textbf{E}_{aug}^* = WarmUpAug(\textbf{E}_{N-1}^*)$ \ELSE~ {$\textbf{E}_{aug}^* = PrototypeAug(\textbf{E}_{N-1}^*, \textbf{E}_{P}^*)$} \ENDIF
        \STATE $\textbf{E}=EMB({f_{\delta}^{ N}}(\textbf{E}_{N-1}))$, $\textbf{E}^*=EMB({f_{\theta}^{ N}}(\textbf{E}_{aug}^*))$
        \STATE $L_{tpl}=TripletMarginLoss(\textbf{E}, \textbf{E}_n, \textbf{E}_p)$
        \STATE $L_{LSC}=ConsistencyLoss(\textbf{E}, \textbf{E}^*)$ \textit{$\rightarrow$LSC loss}
        \STATE $L_{class}=ClassificationLoss(Head(\textbf{E}), label_S)$
        \STATE \textit{BackPropagation(WeightedSum($L_{tpl},L_{class},L_{LSC}$))}
        \ENDFOR
        \IF {\textit{$epoch<N_t-1$}} 
        \STATE $CalculatePrototypes(D_{train})\to Set(\textbf{E}_{N-1})\to PMB$
        \ENDIF
        \ENDFOR
    \end{algorithmic}
\end{algorithm}

\noindent{\textbf{Mixed Attention Fusion Mechanism (MAFM).}}
The MAFM is depicted on the upper right corner of Figure~\ref{fig:Fig2}, which is designed for a further step of aggregation, while considering layer normalization ($LN$), averaged skip connection, and path drop out. First, the attended embedding $\mathbf{E}_{att}$ is obtained through Eq.~(\ref{eq:mixed_fusion}):
\begin{equation}\label{eq:mixed_fusion}
    \mathbf{E}_{att} = MF(LN(\mathbf{E}_{patch,j}),~LN(\mathbf{E}_{patch,v}),~LN(\mathbf{E}_{patch,j})).
\end{equation}
As shown in Figure~\ref{fig:Fig2}, the original patch embeddings $\mathbf{E}_{patch,j}$, $\mathbf{E}_{patch,v}$, and $\mathbf{E}_{patch,b}$ are firstly averaged and then added with the path-dropped attended embedding $\mathbf{E}_{att}$ to have $\mathbf{E}_{asn}$, an embedding after averaging ($AVG$) and applying an skip connection, as depicted in Eq.~(\ref{eq:average}):
\begin{equation}\label{eq:average}
    \mathbf{E}_{asn} = AVG(\mathbf{E}_{patch,j}, \mathbf{E}_{patch,v}, \mathbf{E}_{patch,b}) + DP(\mathbf{E}_{att}),
\end{equation}
where $DP$ indicates the drop path operation. Then, the final mixed resulted embedding $\mathbf{E}_{mixed}$ is obtained via Eq.~(\ref{eq:drop_path}):
\begin{equation}\label{eq:drop_path}
    \mathbf{E}_{mixed} = DP(MLP(LN(\mathbf{E}_{asn}))) + \mathbf{E}_{asn}.
\end{equation}
Finally, the resulted mixed embedding is further fed into the stack of transformer blocks.

\subsection{Prototype-based Latent Space Consistency Loss}
\label{sec:prototype_loss}
To learn data-efficient one-shot action recognition, we put forward a Latent Space Consistency (LSC) loss, which encourages the consistency of the embeddings learned through the main branch and an additional prototype-based feature augmentation branch by cosine similarity loss, as illustrated in Alg.~\ref{algorithm1}.
The intention behind the design of LSC loss is to increase the robustness of the model by forcing the model to learn consistent embeddings even under the disruption of the feature-level augmentation, whose capability against occlusion is verified through our experiments.
We build on top of a recent feature augmentation approach from semi-supervised learning~\cite{kuo2020featmatch}, but additionally propose a warm-up self-augmentation phase and certain architecture changes, which have proven to be effective in improving both the accuracy and the robustness of the model.

\noindent\textbf{Estimating action category prototypes.} 
For the auxiliary branch augmentations at feature-level, we draw inspiration from FeatMatch~\cite{kuo2020featmatch}, a recent method for semi-supervised image classification, where a learned weighted combined category-specific prototypes is used to enhance the intermediate features when referring to feature-level augmentations.
Specifically, for each data-rich action category $l_i \in C_{base}$, we iteratively estimate its prototype in the latent space as the center of all training set embeddings of the specific action (we use the embeddings after the $N{-}1$ block if $N$ is our number of transformer blocks).
Note, that unlike FeatMatch, we use the centers of the data-rich base categories available during training (while clustering is used in semi-supervised learning due to absence of labels). 
Every epoch, these action category prototypes are iteratively updated and stored into a fixed-sized vector by category-wise mean average, which we refer to as the \textit{Prototype Memory Bank} (PMB).
These action category prototypes are then used for feature augmentations in order to estimate the consistency cost. 

\noindent{\textbf{Prototype-based feature enhancement with self-augmentation warm-up.}}
Leveraging prototype-based augmentation in context of one-shot learning requires further conceptual changes. 
First, since the prototypes indeed correspond to actual action categories from $C_{base}$ (\textit{i.e.} only one of the current training categories is correct), we first apply Softmax normalization across the channel dimension for prototypes vector $\textbf{E}_{P}^{*}$ and then refine it with the feature $\textbf{E}_{N-1}^{*}$ and project it into an embedding space as $\textbf{E}_{r, N-1}^{*}{=}g_{\mu}^{2}(Softmax(\textbf{E}_{P}^{*}) \cdot \textbf{E}_{N-1}^{*})$, where N indicates the total number of the transformer stage blocks. $\textbf{E}_{N-1}^{*}$ is obtained through $\textbf{E}_{N-1}^*={f_{\theta}^{1 \to N-1}}(\textbf{E}_{patch})$, where $f_{\theta}^{i}$ indicates the $i$-th transformer stage block for the auxiliary branch and $E_{patch}$ is the mixed fused patch embedding $E_{mixed}$.
At the same time, the feature $\textbf{E}_{N-1}^{*}$ is also projected as $\textbf{E}_{l, N-1}^{*} {=} g_{\mu}^{1}(\textbf{E}_{N-1}^{*})$. Then, the attention weight $\textbf{W}$ is calculated as $\textbf{W} {=} Softmax({\textbf{E}_{l, N-1}^{*}}^{T} \textbf{E}_{r,N-1}^{*})$. After aggregating the information coming from prototypes vector $\textbf{E}_{P}^{*}$ to the original feature $\textbf{E}_{N-1}^{*}$ as depicted in Eq.~(\ref{eq:aggregation}): 
\begin{equation}
\label{eq:aggregation}
  \mathbf{E}_{agg, N-1}^{*}=g_{\mu}^{3}([\mathbf{W}\mathbf{E}_{r, N-1}^{*}, \mathbf{E}_{l, N-1}^{*}]),
\end{equation}
the final augmented feature $\textbf{E}_{aug}^*$ is then obtained through a residual connection with the original feature $\textbf{E}_{N-1}^*$ by $\textbf{E}_{aug}^* {=} ReLU(\textbf{E}_{N-1}^*{+} \textbf{E}_{agg, N-1}^*)$, where $g_{\mu}^1$ and $g_{\mu}^2$ indicate two fully-connected (fc) layers (no weight sharing), and $g_{\mu}^3$ indicates a stack of two fc layers with ReLU in between. $[\cdot]$ denotes concatenation. 
\begin{figure}[t]
\begin{center}
\includegraphics[width = 0.4\textwidth]{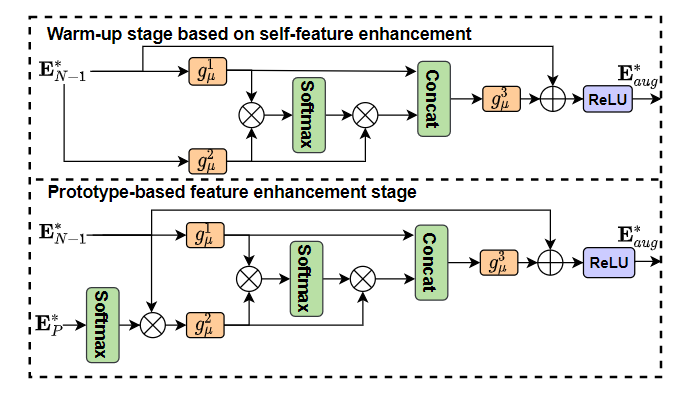}
\end{center}
\caption{Overview of the self-augmentation at feature-level leveraged in the auxiliary branch of the proposed method. During the warm-up phase (top), the feature itself is used as the basis to compute attention masks used to self-augment the feature. At the later stage, we use action-specific prototypes Softmax-normalized along the channel dimension in order to augment the embedding (bottom). 
}
\label{fig:augmentation}
\end{figure}

As in our case the prototypes are linked to true action categories from $C_{base}$ (in contrast to unsupervised clustering necessary in self-supervised tasks), using centers of the assigned categories in the early training epochs would be unreliable. To alleviate this issue, we introduce an additional \textit{warm-up phase}. The key idea is to leverage self-augmentation instead of prototype-based augmentation until certain level of convergence is reached. At earlier stages, we therefore replace the attended prototype representation with the embedding $\textbf{E}_{N-1}^*$. Figure~\ref{fig:augmentation} illustrates the difference between the self-augmentation warm-up phase (top) and the prototype-based augmentation (bottom). Then, we switch to the phase at the bottom of Figure~\ref{fig:augmentation}, while leveraging zero prototypes to achieve decenterization for a fixed number of epochs before using the class-agnostic prototype to do the feature-level prototype-based augmentation.
The final augmented embeddings $E^*$ can be obtained through $\textbf{E}^*=EMB({f_{\theta}^{ N}}(\textbf{E}_{aug}^*))$, where EMB indicates the multi-layer perceptron-based embedding generation layers. The final embedding from the main branch is obtained through $\textbf{E}=EMB(f_{\delta}^{ N}(\textbf{E}_{N-1}))$. The embedding $\textbf{E}_{N-1}$ from the main branch is obtained through  $\textbf{E}_{N-1} = f_{\delta}^{1 \to N-1}(\textbf{E}_{patch})$, where $f_{\delta}^{i}$ indicates the $i$-th transformer stage block of the main branch.
After the acquisition of the embeddings from the main branch $\mathbf{E}$ and the augmented embeddings from the auxiliary branch $\mathbf{E^{\star}}$, the LSC loss is computed as $L_{LSC}=1{-}cos(\mathbf{E},\mathbf{E^{\star}})$.

\subsection{Deep Metric Learning Loss and Classification Loss}
\mypar{Triplet margin loss} To harvest more discriminative embeddings, a triplet margin loss is leveraged in our training pipeline.
Assuming $\textbf{a}_i$, $\textbf{p}_i$, and $\textbf{n}_i$ denote the $i$-th selected anchor, the corresponding positive anchor, and the corresponding negative anchor in the latent space, where the positive anchor shares the same class with the selected anchor and the negative anchor has a different class compared to the selected anchor.
The triplet margin loss is achieved through decreasing the distance between the selected anchor and the positive anchor while increasing the distance between the selected anchor and the negative anchor as depicted in Eq.~(\ref{eq:tpl}):
\begin{equation}
\label{eq:tpl}
    L_{TPL} =\sum_{i=1}^{B} max\{D(\textbf{a}_i,\textbf{n}_i)-D(\textbf{a}_i,\textbf{p}_i) + \sigma, 0\}/B,
\end{equation}
where $\sigma$ denotes the predefined margin and $D(\cdot)$ denotes the pairwise distance. Assuming the pairwise distance between $\textbf{a}$ and $\textbf{n}$ is desired, the pairwise distance can be calculated as $D(\textbf{a},\textbf{n}) = 	\left \| \textbf{a}-\textbf{n} + \epsilon\right \|^2$, where $\epsilon$ keeps as $1e^{-6}$ and $B$ denotes the batch size.

\mypar{Classification loss}
A cross-entropy loss is employed for the supervision of the training to ensure the classifiable capability of the learned embeddings in the latent space. Assuming $\textbf{p}_i$ denotes the prediction of the classifier of the model and $\textbf{y}_i$ denotes the label for the $i$-th sample inside a batch, the cross-entropy loss can be obtained through Eq.~(\ref{eq:cross_entropy}):
\begin{equation}
\label{eq:cross_entropy}
    L_{CLS} =\sum_{i=1}^{B} -\textbf{y}_{i}log(\textbf{p}_{i})+(1-
 \textbf{y}_{i})log(1-\textbf{p}_{i}))/B
\end{equation}

\section{Experiments}
\subsection{Dataset Introduction}
We perform comprehensive studies for the SOAR on three challenging datasets: NTU-60~\cite{shahroudy2016ntu}, NTU-120~\cite{liu2019ntu} and Toyota Smart Home~\cite{Das_2019_ICCV}.
We follow the SOAR protocol in NTU-120 and formulate the evaluation protocols of Toyota Smart Home and NTU-60 to suit our data-scarce representation learning task. Additionally, we propose the occluded SOAR benchmarks for the first time building on top of these three datasets (see Sec. \ref{sec:benchmark}). The NTU-120/NTU-60/Toyota Smart Home benchmarks feature $100/48/24$ data-rich training categories and $20/12/7$ data-scarce test categories respectively for one reference per unseen category. The protocols and occluded datasets will be publicly available in our benchmark.
\begin{table}
\centering
\caption{Experiments for SOAR without occlusion on NTU-120~\cite{liu2019ntu}.}
\label{tab:ntu_120_normal}
\scalebox{0.85}{\begin{tabular}{lcccc} 
\toprule
\textbf{Encoder} & \multicolumn{1}{l}{\textbf{Accuracy}} & \multicolumn{1}{l}{\textbf{F1}} & \multicolumn{1}{l}{\textbf{Precision}} & \multicolumn{1}{l}{\textbf{Recall}} \\ 
\midrule
\multicolumn{5}{l}{\textbf{Previously Published Approaches}} \\ 
\midrule
AN$^\dagger$~\cite{liu2017global} & 41.0 & - & - & - \\
FC$^\dagger$~\cite{liu2017global} & 42.1 & - & - & - \\
AP$^\dagger$~\cite{liu2017global} & 42.9 & - & - & - \\
APSR~\cite{liu2017global} & 45.3 & - & - & - \\
TCN-OneShot~\cite{sabater2021one} & 46.3 & - & - & - \\
SL-DML~\cite{memmesheimer2021sl} & 50.9 & - & - & - \\
Skeleton-DML~\cite{memmesheimer2020skeleton_dml} & 54.2 & - & - & - \\ 
\midrule
\multicolumn{5}{l}{\textbf{CNN-based Encoder Optimized by DML}} \\ 
\midrule
SL-DML (AlexNet~\cite{krizhevsky2014one}) & 40.33 & 39.14 & 42.42 & 40.35 \\
SL-DML (SqueezeNet~\cite{iandola2016squeezenet}) & 42.55 & 40.52 & 41.88 & 42.51 \\
SL-DML (ResNet18~\cite{he2015deep}) & 49.19 & 47.54 & 49.80 & 49.23 \\ 
\midrule
\multicolumn{5}{l}{\textbf{Transformer-based Encoder Optimized with DML (Ours)}} \\ 
\midrule
SL-DML (CaiT~\cite{touvron2021cait}) & 47.86 & 47.53 & 50.06 & 47.94 \\
SL-DML (ViT~\cite{dosovitskiy2020vit}) & 48.45 & 47.40 & 48.59 & 48.52 \\
SL-DML (Twins~\cite{chu2021twins}) & 49.00 & 48.04 & 49.30 & 49.06 \\
SL-DML (ResT~\cite{zhang2021rest}) & 52.58 & 51.86 & 53.99 & 52.61 \\
SL-DML (Swin~\cite{liu2021swin}) & 53.13 & 52.09 & 53.48 & 53.16 \\
SL-DML (LeViT~\cite{graham2021levit}) & 53.19 & 52.22 & 53.85 & 53.29 \\ 
\midrule
\multicolumn{5}{l}{\textbf{Our Proposed and Extended Approaches (Ours)}} \\ 
\midrule
SL-DML (LeViT) + LSC & 55.94 & 54.29 & 55.80 & 56.04 \\
Trans4SOAR (Small) & 56.27 & \textbf{56.43} & \textbf{58.59} & 56.32 \\
Trans4SOAR (Base) & \textbf{57.05} & 55.90 & 57.26 & \textbf{57.12} \\
\bottomrule
\end{tabular}}
\end{table}

\begin{table*}[t]
\centering
\caption{Experiments on the NTU-60~\cite{shahroudy2016ntu} for SOAR considering the scenarios (a) without occlusion, (b) with realistic occlusion (RE) and (c) with random occlusion (RA).}
\label{tab:ntu_60_all}
\scalebox{0.8}{\begin{tabular}{lllllllllllll} 
\toprule
\multirow{2}{*}{\textbf{Encoder}} & \multicolumn{4}{c}{\textbf{(a) Without Occlusion}} & \multicolumn{4}{c}{\textbf{(b) With RE}} & \multicolumn{4}{c}{\textbf{(c) With RA}} \\
 & \multicolumn{1}{c}{\textbf{Acc.}} & \multicolumn{1}{c}{\textbf{F1}} & \multicolumn{1}{c}{\textbf{Prec.}} & \multicolumn{1}{c}{\textbf{Rec.}} & \multicolumn{1}{c}{\textbf{Acc.}} & \multicolumn{1}{c}{\textbf{F1}} & \multicolumn{1}{c}{\textbf{Prec.}} & \multicolumn{1}{c}{\textbf{Rec.}} & \multicolumn{1}{c}{\textbf{Acc.}} & \multicolumn{1}{c}{\textbf{F1}} & \multicolumn{1}{c}{\textbf{Prec.}} & \multicolumn{1}{c}{\textbf{Rec.}} \\ 
\midrule
\multicolumn{13}{l}{\textbf{Previously Published Approaches}} \\ 
\midrule
SL-DML~\cite{memmesheimer2021sl} & 54.82 & 54.31 & 56.72 & 54.65 & 36.90 & 35.86 & 36.59 & 37.05 & 45.28 & 43.13 & 45.00 & 45.42 \\
Skeleton-DML~\cite{memmesheimer2020skeleton_dml} & 55.54 & 50.88 & 53.13 & 51.24 & 42.66 & 40.90 & 41.50 & 42.82 & 60.43 & 59.66 & 61.37 & 60.54 \\ 
\midrule
\multicolumn{13}{l}{\textbf{Transformer-based Encoder Optimized by DML (Ours)}} \\ 
\midrule
SL-DML (Swin~\cite{liu2021swin}) & 56.99 & 56.24 & 58.67 & 56.99 & 51.71 & 50.60 & 52.54 & 51.82 & 64.65 & 63.74 & 66.57 & 64.77 \\
SL-DML (LeViT~\cite{graham2021levit}) & 64.45 & 64.17 & 66.35 & 64.47 & 52.72 & 52.19 & 54.90 & 52.86 & 56.73 & 55.89 & 57.57 & 56.85 \\ 
\midrule
\multicolumn{13}{l}{\textbf{Our Extended and Evaluated Approached (Ours)}} \\ 
\midrule
SL-DML (LeViT) + LSC & 67.67 & 67.87 & 68.74 & 67.67 & 53.79 & 52.76 & 54.18 & 53.88 & 60.78 & 58.75 & 59.97 & 60.90 \\
Trans4SOAR (Small) & 69.74 & 70.52 & 72.45 & 69.82 & 56.84 & 55.84 & 58.27 & 56.98 & 67.90 & 67.32 & 68.94 & 68.01 \\
Trans4SOAR (Base) & \textbf{74.19} & \textbf{74.34} & \textbf{75.91} & \textbf{74.20 } & \textbf{59.28} & \textbf{58.96} & \textbf{59.91} & \textbf{59.40} & \textbf{72.59} & \textbf{71.82} & \textbf{73.89} & \textbf{72.66} \\
\bottomrule
\end{tabular}}
\end{table*}
\begin{table*}[t]
\centering
\caption{Experiments on the Toyota Smart Home~\cite{Das_2019_ICCV} for SOAR (a) without occlusion, (b) with realistic occlusion (RE) and (c) with random occlusion (RA).}
\vskip-1ex
\label{tab:tyt_all}
\scalebox{0.75}{\begin{tabular}{lllllllllllllll} 
\toprule
\multirow{2}{*}{\textbf{Encoder}} & \multicolumn{4}{c}{\textbf{(a) Without Occlusion}} && \multicolumn{4}{c}{\textbf{(b) With RE}} && \multicolumn{4}{c}{\textbf{(c) With RA}} \\
 & \multicolumn{1}{c}{\textbf{Acc.}} & \multicolumn{1}{c}{\textbf{F1}} & \multicolumn{1}{c}{\textbf{Prec.}} & \multicolumn{1}{c}{\textbf{Rec.}} && \multicolumn{1}{c}{\textbf{Acc.}} & \multicolumn{1}{c}{\textbf{F1}} & \multicolumn{1}{c}{\textbf{Prec.}} & \multicolumn{1}{c}{\textbf{Rec.}} && \multicolumn{1}{c}{\textbf{Acc.}} & \multicolumn{1}{c}{\textbf{F1.}} & \multicolumn{1}{c}{\textbf{Prec.}} & \multicolumn{1}{c}{\textbf{Rec.}} \\ 
\midrule
\multicolumn{13}{l}{\textbf{Previously Published Approaches}} \\ 
\midrule
SL-DML~\cite{memmesheimer2021sl} & 58.98 & 27.15 & 27.64 & 35.00& & 38.93 & 25.16 & 32.93 & 28.48 && 53.79 & 26.28 & 27.24 & 29.67 \\
Skeleton-DML~\cite{memmesheimer2020skeleton_dml} & 47.31 & 18.45 & 18.58 & 23.80 && 47.67 & 24.86 & 27.93 & 27.35 && 48.91 & 21.60 & 25.00 & 21.75 \\ 
\midrule
\multicolumn{13}{l}{\textbf{Transformer-based Encoder Optimized by DML (Ours)}} \\ 
\midrule
SL-DML (Swin~\cite{memmesheimer2021sl}) & 58.76 & 28.83 & 29.17 & 32.34 && 35.43 & 18.48 & 23.24 & 23.80 && 65.50 & 29.20 & 30.78 & 29.69 \\
SL-DML (LeViT~\cite{graham2021levit}) & 62.22 & 31.98 & 37.56 & 35.16 && 38.48 & 22.58 & 27.66 & 24.62 && 61.96 & 26.42 & 28.52 & 29.20 \\ 
\midrule
\multicolumn{13}{l}{\textbf{Our Extended and Evaluated Approached (Ours)}} \\ 
\midrule
SL-DML (LeViT) + LSC & 64.46 & 31.91 & 34.07 & 33.58 && 41.82 & 24.34 & 29.02 & 26.67 && 63.77 & 27.72 & 29.09 & 29.90 \\
Trans4SOAR (Small) & 66.87 & 28.08 & 31.47 & 34.63& & 55.12 & \textbf{26.90} & 29.41 & 30.69 && 68.47 & 28.86 & 29.56 & \textbf{32.25} \\
Trans4SOAR (Base) & \textbf{70.22} & \textbf{33.96} & \textbf{37.81} & \textbf{35.33} && \textbf{60.15} & 25.50 & \textbf{33.12} & \textbf{31.86} & &\textbf{68.91} & \textbf{29.27} & \textbf{34.15} & 31.45 \\
\bottomrule
\end{tabular}}
\end{table*}
\begin{table}[t]
\centering
\caption{Ablation study of LSC and MAFM used in the \textsc{Trans4SOAR} on NTU-60~\cite{shahroudy2016ntu} without occlusion.}
\scalebox{0.63}{
\begin{tabular}{llllllll}
\toprule
\textbf{With LSC}&\textbf{Self-aug. wp}&\textbf{De-centerization}& \textbf{MAFM} &\textbf{Accuracy} & \textbf{F1} & \textbf{Precision} & \textbf{Recall} \\

\midrule
&&&&  64.45 & 64.17 & 66.35 & 64.47\\
\checkmark&\checkmark&\checkmark&&  67.67 & 67.87 & 68.74 & 67.67\\
&&&\checkmark& 71.55 & 71.85 & 73.45 & 71.63\\
\checkmark&&&\checkmark& 72.69 & 72.80 & 74.27 & 72.73 \\
\checkmark&&\checkmark&\checkmark& 73.09 & 73.39 & 74.54 & 73.14\\
\checkmark &\checkmark&\checkmark&\checkmark& \textbf{74.19}&\textbf{74.34} &\textbf{75.91} & \textbf{74.20 }\\
\bottomrule
\end{tabular}}
\label{tab:ntu_120_structure_ablation}
\end{table}

\begin{table}[t]
\caption{A comparison to other encoder architectures.}
\center
\scalebox{0.8}{
\begin{tabular}{lllll}
\toprule
\multicolumn{1}{c}{\textbf{Methods}}& \textbf{Accuracy} & \textbf{F1} & \textbf{Recall} & \textbf{Precision} \\
\midrule
SL-DML (CTR-GCN\cite{chen2021channel}) & 43.92 & 41.38 & 45.21 & 43.89  \\
SL-DML (STTR\cite{plizzari2021skeleton}) & 39.56& 39.45 & 41.92 & 39.58 \\
SL-DML (LeViT) + LSC &55.94 & 54.29 & 55.80 & 56.04\\
Trans4SOAR (Small) & 56.27 & \textbf{56.43} & \textbf{58.59} & 56.32 \\
Trans4SOAR (Base) & \textbf{57.05} & 55.90 & 57.26 & \textbf{57.12} \\
\bottomrule
\end{tabular}}
\label{tab:ntu_120_other_experiments}
\end{table}

\begin{table}[t]
\centering
\caption{Experiments for SOAR without occlusion on NTU-120~\cite{liu2019ntu} under Gaussian noise disruption.}
\label{tab:noise}
\scalebox{0.72}{\begin{tabular}{lllllllll} 
\toprule
\multicolumn{5}{l}{\textbf{Gaussian Noise~ ~ ~~~~~ ~ ~~$\sigma=0.1, \mu = 0$}} & \multicolumn{4}{l}{\textbf{\textbf{~$\sigma=0.05, \mu = 0$}}} \\
\textbf{Encoder} & \multicolumn{1}{c}{\textbf{Acc.}} & \multicolumn{1}{c}{\textbf{F1}} & \multicolumn{1}{c}{\textbf{Prec.}} & \multicolumn{1}{c}{\textbf{Rec.}}  & \multicolumn{1}{c}{\textbf{Acc.}} & \multicolumn{1}{c}{\textbf{F1}} & \multicolumn{1}{c}{\textbf{Prec.}} & \multicolumn{1}{c}{\textbf{Rec.}} \\ 
\midrule
SL-DML~\cite{memmesheimer2021sl} &  21.42 & 11.83 & 8.50 & 21.71  & 21.76 & 12.23 & 8.70 & 21.86 \\
SL-DML (LeViT) &  22.31 & 12.32 & 8.79 & 22.40   & 21.97 & 12.82 & 9.69 & 22.07 \\
SL-DML (LeViT) + LSC &  52.54 & 51.16 & 51.61 & 52.65   & 51.91 & 50.08 & 51.67 & 52.01 \\
Trans4SOAR &  \textbf{53.09} & \textbf{51.89} & \textbf{53.05} & \textbf{53.15}   & \textbf{\textbf{54.74}} & \textbf{\textbf{54.65}} & \textbf{\textbf{56.33}} & \textbf{\textbf{54.83}} \\
\bottomrule
\end{tabular}}
\end{table}

\begin{table}[t]
\centering

\caption{Experiments regarding realistic synthesized occlusion (a) and random occlusion (b) for SOAR on NTU-120~\cite{liu2019ntu}.}
\label{tab:ntu_120_occ}
\scalebox{0.75}{\begin{tabular}{lccccllll} 
\toprule
\multirow{2}{*}{\textbf{Encoder}} & \multicolumn{4}{c}{\textbf{(a) With RE}} & \multicolumn{4}{c}{\textbf{(b) With RA}} \\
 & \multicolumn{1}{l}{\textbf{Acc.}} & \multicolumn{1}{l}{\textbf{F1}} & \multicolumn{1}{l}{\textbf{Prec.}} & \multicolumn{1}{l}{\textbf{Rec.}} & \textbf{Acc.} & \textbf{F1} & \textbf{Prec.} & \textbf{Rec.} \\ 
\midrule
SL-DML~\cite{memmesheimer2021sl}  & 39.82 & 37.85 & 39.32 & 39.86& 42.53 & 42.24 & 44.79 & 42.56 \\
Skeleton-DML~\cite{memmesheimer2020skeleton_dml} & 49.21 & 46.82 & 48.10 & 49.18 & 35.15 & 32.59 & 34.29 & 35.22 \\
SL-DML (LeViT~\cite{graham2021levit})  & 44.22 & 42.29 & 44.20 & 44.31& 35.00 & 33.24 & 41.45 & 35.10 \\
SL-DML (Swin~\cite{liu2021swin}) & 47.19 & 45.64 & 46.78 & 47.29 & 47.19 & 45.64 & 46.78 & 47.29 \\ 
\midrule
SL-DML (LeViT) + LSC & 48.28 & 46.03 & 47.58 & 48.31& 38.04 & 35.93 & 37.87 & 38.11  \\
Trans4SOAR (Small) & 51.64 & \textbf{50.47} & 52.36 & 51.70& \textbf{53.27} & 51.33 & 53.80 & \textbf{53.35}  \\
Trans4SOAR (Base) & \textbf{52.35} & 48.79 & \textbf{52.87} & \textbf{52.43}& 53.17 & \textbf{52.89} & \textbf{54.50} & 53.21  \\
\bottomrule
\end{tabular}}
\end{table}
\begin{table}[t]
\centering
\caption{Experiments with different realistic synthesized occlusion ratio on the NTU-60~\cite{shahroudy2016ntu} for the SOAR.}
\label{tab:ntu_60_different_realistic ratio}
\scalebox{0.75}{
\begin{tabular}{l|c|llll} 
\toprule
\multicolumn{1}{l}{\textbf{Model}} & \multicolumn{1}{l}{\textbf{RE\_Range}} & \textbf{Accuracy} & \textbf{F1} & \textbf{Precision} & \textbf{Recall} \\ 
\midrule
SL-DML~\cite{memmesheimer2021sl} & \multirow{6}{*}{0.05-0.2} & 36.90 & 35.86 & 36.59 & 37.05 \\
Skeleton-DML~\cite{memmesheimer2020skeleton_dml} & &35.15 & 32.59& 34.29& 35.22\\
SL-DML (LeViT~\cite{graham2021levit}) & & 52.72 & 52.19 & 54.90 & 52.86 \\
SL-DML (LeViT) + LSC &  & 53.79 & 52.76 & 54.18 & 53.88 \\
Trans4SOAR (Small) & &  56.84& 55.84 & 58.27 & 56.98 \\
Trans4SOAR (Base) &  & \textbf{59.28} & \textbf{58.96} & \textbf{59.91} & \textbf{59.40} \\ 
\midrule
SL-DML~\cite{memmesheimer2021sl} & \multirow{6}{*}{0.05-0.35} & 39.26 & 38.71 &39.59 & 39.43 \\
Skeleton-DML ~\cite{memmesheimer2020skeleton_dml}&   & 38.52 & 38.74 & 39.23 & 38.64   \\
SL-DML (LeViT~\cite{graham2021levit}) && 53.17  & 52.52 & 54.16 & 53.34    \\
SL-DML (LeViT) + LSC &  & 53.58 & 52.75 & 54.07 & 53.77 \\
Trans4SOAR (Small) & & \textbf{61.69} & \textbf{61.60} & \textbf{64.01}&\textbf{61.81} \\
Trans4SOAR (Base) &  & 58.27 & 56.63 & 58.81 & 58.40 \\ 
\midrule
SL-DML~\cite{memmesheimer2021sl} & \multirow{6}{*}{0.05-0.5} & 34.89 & 32.63 & 31.85 & 35.07 \\
Skeleton-DML~\cite{memmesheimer2020skeleton_dml}&   &  42.83& 42.33 & 42.46 & 42.93   \\
SL-DML (LeViT~\cite{graham2021levit}) & & 54.84 & 54.07&57.06 &54.99 \\

SL-DML (LeViT) + LSC &  & 55.07 & 55.01& 57.56& 55.21 \\
Trans4SOAR (Small) & & \textbf{59.59} & \textbf{59.21}& 59.49& \textbf{59.70} \\
Trans4SOAR (Base)&  & 57.52 & 57.21 & \textbf{59.61} & 57.64 \\ 
\bottomrule
\end{tabular}}
\end{table}

\begin{table}[t]
\centering
\vskip-2.5ex
\caption{Experiments for different fusion techniques on  NTU60~\cite{shahroudy2016ntu} under different occlusion scenarios.}
\vskip-1ex
\label{tab:ntu_60_fusion_comparison}
\scalebox{0.7}{\begin{tabular}{llllll}
\toprule
\textbf{Fusion Method}& \textbf{OCC} & \textbf{Accuracy} & \textbf{F1} & \textbf{Precision} & \textbf{Recall} \\ 
\midrule
Single (Joints) & RE & 53.79 & 52.76 & 54.18 & 53.88 \\
Single (Bones) & RE & 54.22 & 53.73 & 54.86 & 54.33 \\
Single (Velocities) & RE & 56.93 & 56.10 & 57.97 & 57.03  \\
Addition & RE & 56.37 & 54.48 & 55.68 & 56.51 \\
Multiplication & RE & 53.35 & 51.91 & 53.69 & 53.50 \\
Concatenation & RE & 58.61 & 57.21 & 57.63 & 58.73 \\
Late Fusion & RE & 56.93 & 56.10 & 57.97 & 57.03 \\
\midrule
MAFM & RE & \textbf{59.28} & \textbf{58.96} & \textbf{59.91} & \textbf{59.40} \\ 
\midrule
Single (Joints) & RA & 60.78 & 58.75 & 59.97 & 60.90 \\
Single (Bones) & RA & 55.15 & 53.56 & 56.63 & 54.16 \\
Single (Velocities) & RA & 33.15 & 30.54 & 29.67 & 33.82  \\
Addition & RA & 65.09 & 65.03 & 66.36 & 65.18 \\
Multiplication & RA & 67.54 & 67.51 & 68.65 & 67.63 \\
Concatenation & RA & 68.05 & 68.54 & 70.90 & 68.13 \\
Late Fusion & RA & 71.16 & 71.58 & 73.16 & 71.22 \\
\midrule
MAFM & RA & \textbf{72.59}&\textbf{71.82} &\textbf{73.89} & \textbf{72.66 }\\
\midrule
Single (Joints) & N & 67.67 & 67.87 & 68.74 & 67.67 \\
Single (Bones) & N &61.45  & 61.44 & 63.50 & 61.57 \\
Single (Velocities) & N & 49.74 & 50.08 & 51.31 & 49.89 \\
Addition & N & 67.05 & 66.88 & 68.09 & 67.12 \\
Multiplication  & N & 64.63 & 65.05 & 66.34 & 64.75 \\
Concatenation  & N & 67.75 & 67.79 & 69.56 & 67.86 \\
Late Fusion  & N & 57.15 & 56.52 & 57.57 & 57.26 \\
\midrule
MAFM & N & 
 \textbf{74.19}&\textbf{74.34} &\textbf{75.91} & \textbf{74.20 }\\
\bottomrule
\end{tabular}}
\end{table}

\begin{table}[t]
\centering
\caption{Experiments regarding different random occlusion ratio on the NTU-60~\cite{shahroudy2016ntu} for the SOAR.}
\label{tab:ntu_60_different_random_ratio}
\scalebox{0.7}{
\begin{tabular}{l|c|llll} 
\toprule
\multicolumn{1}{l}{\textbf{Model}} & \multicolumn{1}{l}{\textbf{RA\_ratio}} & \textbf{Accuracy} & \textbf{F1} & \textbf{Precision} & \textbf{Recall} \\ 
\midrule
SL-MDL~\cite{memmesheimer2021sl} & \multirow{6}{*}{0.1} & 45.28 & 43.13 & 45.00 & 45.42 \\
Skeleton-DML~\cite{memmesheimer2020skeleton_dml}& & 60.43 & 59.66 & 61.37 & 60.54   \\
SL-DML (LeViT~\cite{graham2021levit})&  & 56.73 & 55.89 & 57.75 & 56.85  \\
SL-DML (LeViT) + LSC&  & 60.78 & 58.75 & 59.97 & 60.90  \\

Trans4SOAR (Small) && 69.74 & 70.52 & 72.45 & 69.82  \\
Trans4SOAR (Base) && \textbf{72.59}&\textbf{71.82} &\textbf{73.89} & \textbf{72.66 }\\
\midrule
SL-DML~\cite{memmesheimer2021sl} & \multirow{6}{*}{0.3} &  46.39 & 42.82 & 46.69 & 46.54   \\
Skeleton-DML~\cite{memmesheimer2020skeleton_dml} && 58.93 & 56.07 & 58.45 & 59.05   \\
SL-DML (LeViT~\cite{graham2021levit})&  & 46.32 & 43.78 & 43.94 & 46.40 \\
SL-DML (LeViT) + LSC &  & 47.82 & 45.02 & 48.41 & 47.91 \\
Trans4SOAR (Small) &  & 66.57 & 66.26 & 67.94 & 66.65 \\
Trans4SOAR (Base) &  & \textbf{72.39} & \textbf{72.81} & \textbf{74.68} & \textbf{72.43} \\ 
\midrule
SL-DML~\cite{memmesheimer2021sl} & \multirow{6}{*}{0.5} & 43.44 & 38.46 & 41.30 & 43.57 \\
Skeleton-DML~\cite{memmesheimer2020skeleton_dml} && 44.69 & 41.89 & 45.74 &  44.79  \\
SL-DML (LeViT~\cite{graham2021levit})&  & 35.77 & 32.56 & 36.22 & 35.94 \\
SL-DML (LeViT) + LSC&  & 40.53 & 37.38 & 38.33  & 40.59 \\
Trans4SOAR (Small) &  & 52.92 & 50.78 & 55.13 & 53.02 \\
Trans4SOAR (Base) & &\textbf{54.82} & \textbf{55.01} & \textbf{58.01} & \textbf{54.93}  \\
\bottomrule
\end{tabular}}
\end{table}

\begin{table}[t]
\centering
\vskip-2ex
\caption{Experiments for reference w/ or w/o occlusions on NTU-60~\cite{shahroudy2016ntu}.}
\label{tab:ntu_60_reference_ablation}
\vskip-1ex
\scalebox{0.7}{
\begin{tabular}{l|c|c|llll} 
\toprule
\multicolumn{1}{l}{\textbf{Model}} & \multicolumn{1}{l}{\textbf{OCC}} & \multicolumn{1}{l}{\textbf{OCCVal}} & \textbf{Accuracy} & \textbf{F1} & \textbf{Precision} & \textbf{Recall} \\ 
\midrule
SL-MDL~\cite{memmesheimer2021sl} & \multirow{6}{*}{RA} & \multirow{6}{*}{T} & 48.74 & 46.46 & 47.45 & 48.88  \\
Skeleton-DML~\cite{memmesheimer2020skeleton_dml} & & &49.30 & 48.57& 49.62& 49.45 \\
SL-DML (LeViT~\cite{graham2021levit}) & & & 53.47 & 52.35 & 54.94 & 53.63 \\

SL-DML (LeViT) + LSC & & & 53.57 & 53.73 & 56.55 & 53.72  \\
Trans4SOAR (Small)&  &  & \textbf{72.16} & \textbf{72.42} & 73.67 & \textbf{72.23} \\
Trans4SOAR (Base) &  &  &  71.59 & 72.22 & \textbf{73.95} & 71.67 \\ 
\midrule
SL-DML~\cite{memmesheimer2021sl} & \multirow{6}{*}{RA} & \multirow{6}{*}{F} & 45.28 & 43.13& 45.00& 45.42 \\
Skeleton-DML~\cite{memmesheimer2020skeleton_dml} & & & 60.43 & 59.66&61.37 &60.54 \\
SL-DML (LeViT~\cite{graham2021levit})& && 56.73 & 55.89 & 57.57& 56.85 \\

SL-DML (LeViT) + LSC &&  & 60.78 & 58.75 & 59.97 & 60.90   \\
Trans4SOAR (Small)& && 67.90 & 67.32 & 68.94 & 68.01 \\

Trans4SOAR (Base) &  &&\textbf{72.59}&\textbf{71.82} &\textbf{73.89} & \textbf{72.66 }\\
\midrule
SL-DML~\cite{memmesheimer2021sl} & \multirow{6}{*}{RE} & \multirow{6}{*}{T} & 36.90 & 35.86 & 36.59 & 37.05 \\
Skeleton-DML~\cite{memmesheimer2020skeleton_dml} & & & 42.66 & 40.90& 41.50& 42.82\\
SL-DML (LeViT~\cite{graham2021levit}) & & & 52.72 & 52.19& 54.90& 52.86\\

SL-DML (LeViT) + LSC &  &  & 53.79 & 52.76 & 54.18 & 53.88 \\
Trans4SOAR (Small)&&& 56.84& 55.84 & 58.27 & 56.98 \\

Trans4SOAR (Base) &&& \textbf{59.28} & \textbf{58.96} & \textbf{59.91} &\textbf{59.40} \\ 
\midrule
SL-DML~\cite{memmesheimer2021sl} & \multirow{4}{*}{RE} & \multirow{4}{*}{F} & 39.51 & 39.64 & 40.82 & 39.64  \\
Skeleton-DML~\cite{memmesheimer2020skeleton_dml} & & & 44.29 & 43.10& 44.26& 44.46\\
SL-DML (LeViT~\cite{graham2021levit}) & & & 55.12 & 55.22& 57.51& 55.26\\

SL-DML (LeViT) + LSC & & & 55.07 & 55.01 & 57.56 &55.21 \\
Trans4SOAR (Small)&  &  & 54.37 & 52.97 & 55.08 &54.38  \\

Trans4SOAR (Base) & & & \textbf{58.48} & \textbf{57.10} & \textbf{57.75} & \textbf{58.61}  \\
\bottomrule
\end{tabular}}
\end{table}
\subsection{Implementation Details}
For \textsc{Trans4SOAR} training we set the warm-up phase threshold $N_t = 20$ while using another $10$ epochs for decenterization. We train our model optimized by AdamW~\cite{loshchilov2017decoupled} with Cosine Annealing Scheduler for $50$ epochs and batch size of $32$ using Nvidia A100 GPU with PyTorch 1.8.0 to reproduce the best performance. We use an initial learning rate of $3.5e^{-5}$ with the weights of the three losses, \textit{i.e.}, Triplet Margin Loss ($\sigma=0.2$), Cross Entropy Loss and LSC loss as $1.0$, $0.4$ and $0.1$. Our \textsc{Trans4SOAR} (Small) has $D_{Key}$: $1$, $N_{head}:[2,2,2]$, $H_{dep}:[2,4,4]$ and $C_{dim}:[384,512,512]$ with 23M parameters while \textsc{Trans4SOAR} has $D_{Key}: 32$, $N_{head}$:$[6,9,12]$, $H_{dep}:[4,4,4]$ and $C_{dim}:[384,512,768]$ with 43M parameters, where $D_{Key}$, $N_{head}$, $H_{dep}$ and $C_{dim}$ denote dimension of Key, number of the attention head, number of the basic transformer attention block inside each transformer block and the unified feature dimension inside each transformer block respectively. Both of our approaches have three main transformer blocks.
To ensure that there is no information leakage caused by the data augmentation to the occlusion region, the occlusion is generated before the data augmentation for both the realistic occlusion scenario and the random occlusion scenario.
The protocols and the occlusion benchmarks will be released.
\subsection{Analyses for SOAR Without Occlusion}

\mypar{Performance analyses regarding different components}
As in Table~\ref{tab:ntu_120_normal}, we firstly empirically evaluate the benefits brought by the LSC loss achieved through prototype-based feature augmentation and an additional auxiliary branch.
The baseline we chose is SL-DML~\cite{memmesheimer2021sl}, which has the same data preprocessing technique with our approach.
Specifically, we use the SL-DML pipeline and equip the selected transformer-based architecture, \textit{i.e.}, LeViT~\cite{graham2021levit}, with an additional auxiliary branch for attention-based augmentations via feature-level prototypes. The LSC loss is obtained through the calculation of cosine similarity loss between the embedding from the main branch and the embedding from the auxiliary branch. The aforementioned approach with LSC loss is denoted as SL-DML (LeViT) + LSC compared with SL-DML (LeViT), which replaces CNN by the LeViT in the SL-DML pipeline.
Although LSC loss does not have any influence on the architecture at test-time, it performs surprisingly well for the SOAR task without occlusion. For instance, SL-DML (LeViT) + LSC Loss leads to accuracy gains by $2.75\%$ (NTU-120, Table \ref{tab:ntu_120_normal}), $3.22\%$ (NTU-60, Table~\ref{tab:ntu_60_all} (a)) and $2.24\%$ (Toyota Smart Home, Table \ref{tab:tyt_all} (a)), compared with SL-DML (LeViT), which has overall better performance compared with SL-DML~\cite{memmesheimer2021sl} and Skeleton-DML~\cite{memmesheimer2020skeleton_dml} for the SOAR without occlusion.
We observe the benefits of our LSC loss on NTU-120~\cite{liu2019ntu}, surpassing the previous two approaches, \textit{i.e.}, SL-DML~\cite{memmesheimer2021sl} by ${>}5\%$ and Skeleton-DML~\cite{memmesheimer2020skeleton_dml} by $1.74\%$ (Table \ref{tab:ntu_120_normal}). 
Our ablation experiments regarding the main components of LSC loss are shown in Table \ref{tab:ntu_120_structure_ablation} regarding the last three experiments, showing the importance of the warm-up stage and de-centerization stage which bring a performance improvement by $1.5\%$ compared with LSC loss without both the aforementioned components.
Then, the combination of the LSC loss and the MAFM, mixing three streams of input at patch embedding level, further contributes a remarkable performance gain regarding the SOAR without occlusion compared with the existing state-of-the-art works~\cite{memmesheimer2020skeleton_dml,memmesheimer2020skeleton_dml}. On the NTU-120~\cite{liu2019ntu}, \textsc{Trans4SOAR} (Base) surpasses Skeleton-DML~\cite{memmesheimer2020skeleton_dml} and SL-DML~\cite{memmesheimer2021sl} by $2.85\%$ and $6.15\%$ for accuracy while outperforming SL-DML (LeViT) + LSC by $1.11\%$, indicating an incremental performance enhancement considering the discriminative ability of the learned embedding by using MAFM and LSC loss. We also conduct experiments to showcase the individual performance gain brought by LSC and MAFM in Table~\ref{tab:ntu_120_structure_ablation} regarding the first three experiments.
Furthermore, consistent improvements are achieved by \textsc{Trans4SOAR} in the other two datasets, \textit{e.g.}, NTU-60 in Table~\ref{tab:ntu_60_all} (a) and Toyota Smart Home~\cite{Das_2019_ICCV} in Table~\ref{tab:tyt_all} (a) for the SOAR without occlusion.
The NTU-60~\cite{shahroudy2016ntu} has less training categories than the NTU-120~\cite{liu2019ntu}, thus, it is used to evaluate the generalizability of the leveraged models, which means realizing the SOAR with less a prior knowledge.
In Table~\ref{tab:ntu_60_all} (a), our \textsc{Trans4SOAR} (Base) surpasses SL-DML~\cite{memmesheimer2021sl} and Skeleton-DML~\cite{memmesheimer2020skeleton_dml} by $19.37\%$ and $18.65\%$ for accuracy, indicating that, given less a prior knowledge, \textsc{Trans4SOAR} has better capability to harvest more discriminative representation. 
Furthermore, the Toyota Smart Home~\cite{Das_2019_ICCV} contains 2D skeleton data in image coordinate format, delivering a valuable data format to explore the SOAR task.
In Table~\ref{tab:tyt_all} (a), our \textsc{Trans4SOAR} (Base) undoubtedly shows the best performance over all the previous approaches with large margin.
Observing the other three metrics, \textit{i.e.}, F1-score, precision and recall, since the first two datasets have balanced distributed samples for different categories, theses three terms do not have large difference compared with the accuracy. However, since the action categories on the Toyota Smart Home~\cite{Das_2019_ICCV} is not equal distributed, these three terms are able to showcase whether the true prediction is balanced distributed in the test set or not.
Our \textsc{Trans4SOAR} surpasses all the approaches in terms of all metrics on the investigated datasets.
In order to ablate the effect of different model scales, we construct \textsc{Trans4SOAR} (Small) with only 23M parameters which pursues both light model structure and high accuracy, and achieves second best performance, showcasing that the LSC loss and MAFM are helpful for learning discriminative features via different model variants.
We also conduct experiments in Table~\ref{tab:ntu_120_other_experiments} to compare with graph convolutional approach~\cite{chen2021channel} and skeleton transformer approach~\cite{plizzari2021skeleton}, however the performance of these two encoder architectures for the SOAR task even without occlusion is not satisfied compared with \textsc{Trans4SOAR} and SL-DML (LeViT). 

\begin{table}[t]
\centering
\caption{Experiments for random temporal and spatial occlusion.}
\label{tab:ntu_60_ramdomly_temporal_and_spatial_occ}
\scalebox{0.68}{\begin{tabular}{lllllllll} 
\toprule
\multirow{2}{*}{\textbf{Model}} & \multicolumn{4}{l}{\textbf{(a) Random temporal occlusion}} & \multicolumn{4}{l}{\textbf{(b) Random spatial occlusion}} \\
 & \textbf{Acc.} & \textbf{F1.} & \textbf{Prec.} & \textbf{Rec.} & \textbf{Acc.} & \textbf{F1.} & \textbf{Prec.} & \textbf{Rec.} \\ 
\midrule
\multicolumn{9}{l}{\textbf{Experiments on NTU-120~ with random temporal occlusion.}} \\ 
\midrule
SL-DML~\cite{memmesheimer2021sl} & 38.15 & 34.87 & 38.51 & 38.11 & 38.15 & 35.26 & 36.76 & 38.13 \\
Skeleton-DML~\cite{memmesheimer2020skeleton_dml} & 27.20 & 24.43 & 26.75 & 27.12 & 27.93 & 25.91 & 28.24 & 27.93 \\
Trans4SOAR (Small) & 51.60 & 50.73 & 52.65 & 50.99 & 46.99 & 46.24 & 49.71 & 47.07 \\
Trans4SOAR (Base) & \textbf{54.11} & \textbf{52.93} & \textbf{53.85} & \textbf{54.21} & \textbf{49.43} & \textbf{49.08} & \textbf{51.35} & \textbf{49.48} \\ 
\midrule
\multicolumn{9}{l}{\textbf{Experiments on NTU-60~ with random temporal occlusion.}} \\ 
\midrule
SL-DML~\cite{memmesheimer2021sl} & 58.68 & 58.46 & 60.20 & 58.72 & 52.48 & 50.59 & 54.06 & 52.65 \\
Skeleton-DML~\cite{memmesheimer2020skeleton_dml} & 51.81 & 50.50 & 53.06 & 51.95 & 46.38 & 43.68 & 45.94 & 46.54 \\
Trans4SOAR (Small) & 71.45 & 71.32 & 72.94 & 71.51 & 68.94 & 69.61 & \textbf{71.84} & 69.01 \\
Trans4SOAR (Base) & \textbf{75.01} & \textbf{74.75} & \textbf{75.76} & \textbf{75.06} & \textbf{69.08} & \textbf{69.18} & 71.19 & \textbf{69.14} \\ 
\midrule
\multicolumn{9}{l}{\textbf{Experiments on Toyota Smart Home~ with random temporal occlusion.}} \\ 
\midrule
SL-DML~\cite{memmesheimer2021sl} & 53.36 & 22.97 & 28.17 & 24.58 & 60.36 & 20.10 & 24.52 & 20.89 \\
Skeleton-DML~\cite{memmesheimer2020skeleton_dml} & 53.65 & 23.90 & 31.54 & 25.33 & 41.95 & 26.36 & 32.83 & 27.97 \\
Trans4SOAR (Small) & 63.66 & 29.90 & 31.76 & 34.06 & \textbf{66.76} & 31.76 & 33.14 & \textbf{35.66} \\
Trans4SOAR (Base) & \textbf{68.48} & \textbf{31.11} & \textbf{33.81} & \textbf{33.80} & 64.49 & \textbf{32.43} & \textbf{35.80} & 34.29 \\
\bottomrule
\end{tabular}}
\end{table}

\mypar{Tolerance to noisy inputs}
The quality of the skeleton data is influenced by a variety of factors, such as sensor noise or occlusions. First, a larger gap between the \textsc{Trans4SOAR} and standard DML trained on the Toyota Smart Home~\cite{Das_2019_ICCV} (which is noisier than the more controlled NTU-datasets) hints towards its advantages specifically for imperfect input. To validate if this is the case, we evaluate the model for inputs corrupted by different magnitudes of Gaussian noise and discover a remarkable tolerance of \textsc{Trans4SOAR} (in Table \ref{tab:noise}).
While the prediction quality diminishes for basic DML-based models, the utilizing of LSC loss on the SL-DML (LeViT) is more robust when confronted with unreliable data, which showcases the superiority of the proposed LSC loss against Gaussian noise input. 

In particular, the performance for the SL-DML (LeViT) with the LSC loss falls from $55.94\%$ on clean data to $51.91\%$ for Gaussian noise with $\sigma {=} 0.05$, while this decline is much higher ($53.19\% \rightarrow 21.97\%$) for the SL-DML (LeViT). We attribute this to the extensive learned augmentations at the feature-level taking place in the auxiliary branch while formulating the LSC loss. The LSC loss encourages the model to output similar results if the embedding has been altered, which suits naturally to the use-case of noise disturbances.
Furthermore, \textsc{Trans4SOAR} (Base) surpasses all the other investigated approaches with no doubt by $54.74\%$ and $53.09\%$ in terms of accuracy for the Gaussian noise conditioned by $\sigma {=} 0.05$ and $\sigma {=} 0.1$ respectively.

\subsection{Analyses for REalistic Synthesized Occlusion (RE)}
We conduct experiments regarding RE for NTU-120~\cite{liu2019ntu}, NTU-60~\cite{shahroudy2016ntu} and Toyota Smart Home~\cite{Das_2019_ICCV} in Table~\ref{tab:ntu_120_occ} (a), Table~\ref{tab:ntu_60_all} (b), and Table~\ref{tab:tyt_all} (b) with SNR range $0.05\rightarrow0.2$ and with occlusion on the reference set.
First, the performance of all investigated approaches for the SOAR with RE benchmark is degraded compared to the SOAR without occlusion benchmark, indicating that the proposed RE is very challenging for discriminative representation learning.
In Table~\ref{tab:ntu_120_occ} (a), our \textsc{Trans4SOAR} (Base) shows the best performance by $52.35\%$, $48.79\%$, $52.87\%$ and $52.43\%$ for accuracy, F1-score, precision and recall, indicating that the performance is equally distributed among the investigated classes on the NTU-120~\cite{liu2019ntu}.
The \textsc{Trans4SOAR} (Small) achieves second best performance among all the metrics on NTU-120 with RE by $51.64\%$ for accuracy. Note, that the SL-DML (LeViT) demonstrates worse performances on all the conducted datasets with RE for SOAR. On NTU-120~\cite{liu2019ntu} with RE, the SL-DML (LeViT) approach only has an accuracy of $44.22\%$ which is lower than the Skeleton-DML~\cite{memmesheimer2020skeleton_dml} with an accuracy of $49.21\%$. However, compared with SL-DML (LeViT), SL-DML (LeViT) + LSC loss still has a better accuracy of $48.28\%$, indicating that LeViT architecture is not good at dealing with RE, while LSC loss can alleviate the negative influence.
After the using of the MAFM to form our \textsc{Trans4SOAR} (Base), a superior performance of $52.35\%$ in accuracy shows up, indicating that the disruption issue caused by RE can be well addressed through the triplet stream encoding and MAFM.
These experimental results illustrate the importance of the proposed MAFM on dealing with the disruption brought by the RE through aggregating three different skeleton encoding formats, which contains potential de-occlusion cues, and also show the superiority of our reformulated \textsc{Trans4SOAR} regarding the robustness against the occlusion disruption from the real life compared with the LeViT, on which we build up our \textsc{Trans4SOAR} based.
Similar comparison and analyses could be found on the other two datasets, \textit{i.e.}, NTU-60~\cite{shahroudy2016ntu} in Table~\ref{tab:ntu_60_all} (b) and Toyota Smart Home~\cite{Das_2019_ICCV} in Table~\ref{tab:tyt_all} (b), where the \textsc{Trans4SOAR} (Base) surpasses Skeleton-DML~\cite{memmesheimer2020skeleton_dml} and SL-DML~\cite{memmesheimer2021sl} by $16.62\%$ and $22.38\%$  on NTU-60~\cite{shahroudy2016ntu}, and $12.48\%$ and $21.22\%$ on Toyota Smart Home~\cite{Das_2019_ICCV}, while \textsc{Trans4SOAR} (Small) also shows competitive performances. We also conduct experiments by using different Sigal-to-Noise Ratio (SNR) range for the SOAR with RE as depicted in Table~\ref{tab:ntu_60_different_realistic ratio}, \textsc{Trans4SOAR} shows promising and stable performance $>56\%$ in terms of accuracy considering both \textsc{Trans4SOAR} (Base) and \textsc{Trans4SOAR} (Small) for three SNR ranges, \textit{i.e.}, $0.05-0.2$, $0.05-0.35$ and $0.05-0.5$ on NTU-60~\cite{shahroudy2016ntu}. 
\begin{figure*}[t]
\begin{center}
\includegraphics[width =\textwidth]{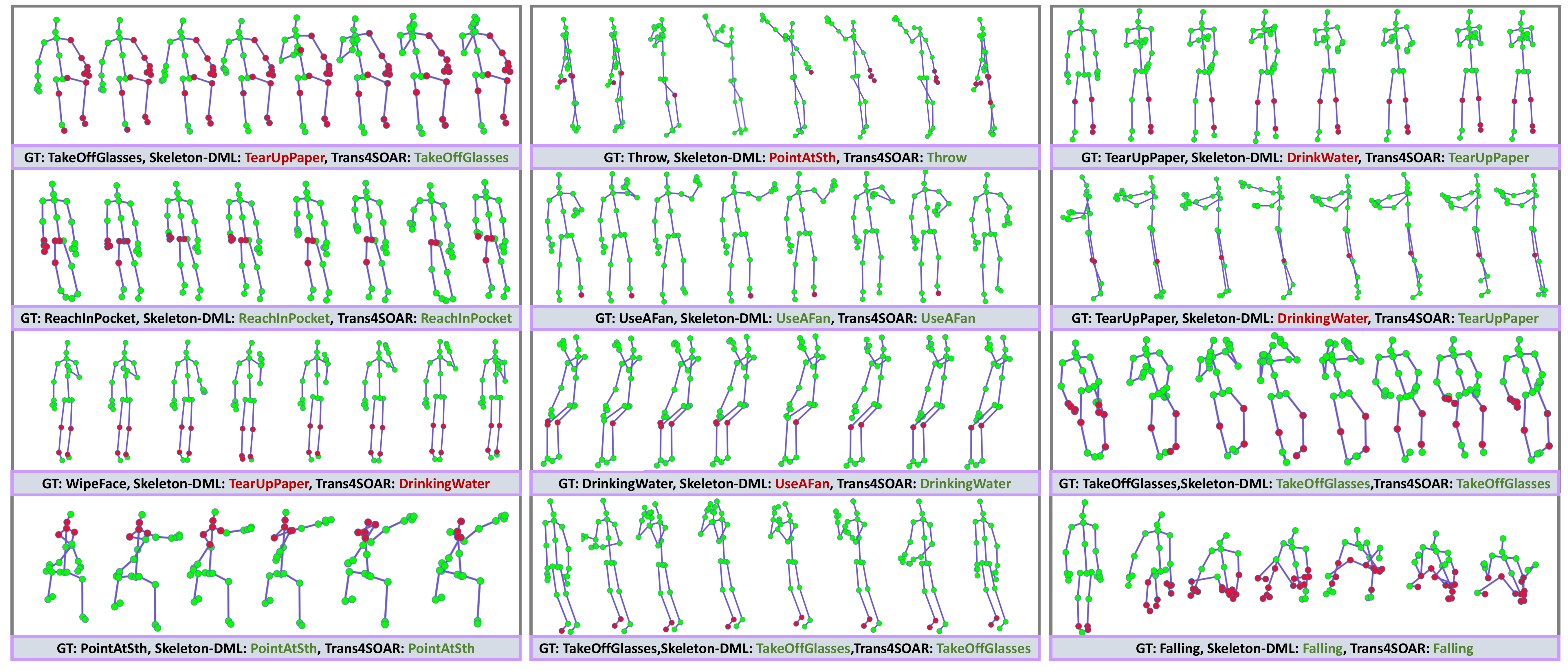}
\end{center}
\caption{An overview of the qualitative experimental results  on NTU-60~\cite{shahroudy2016ntu} with RE for SOAR, where GT indicates the groundtruth and Trans4SOAR indicates the prediction of \textsc{Trans4SOAR}-Base. The true prediction is marked as green, while the false prediction is marked as red. }

\label{fig:qualitative}
\end{figure*}
\begin{figure}[t]
\begin{center}
\includegraphics[width =0.45\textwidth]{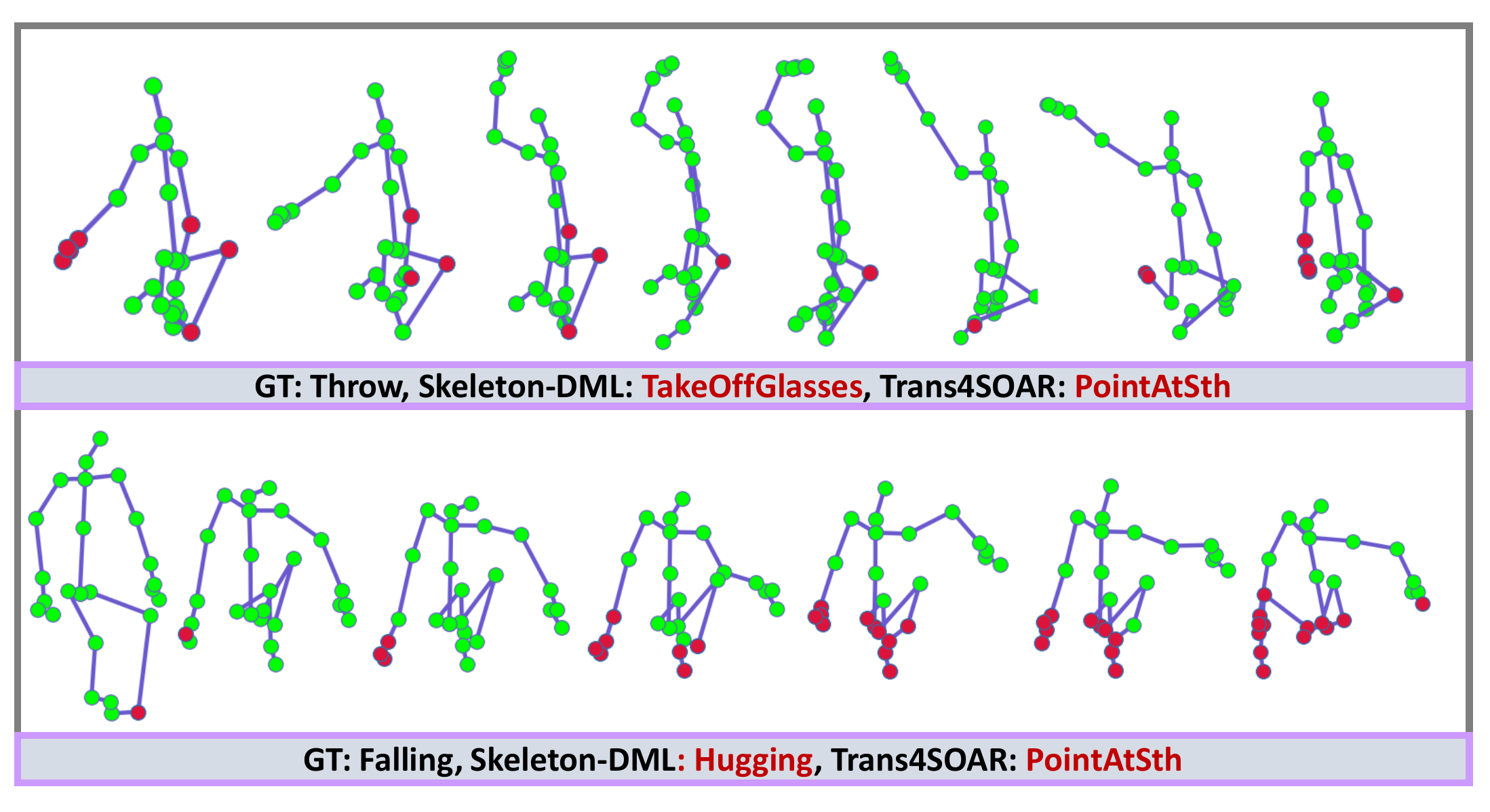}
\end{center}
\caption{Failure case examples for SOAR with RE on NTU-60~\cite{shahroudy2016ntu}.}
\label{fig:failure_case}
\end{figure}
\begin{figure}[t]
\begin{center}
\includegraphics[width =0.4\textwidth]{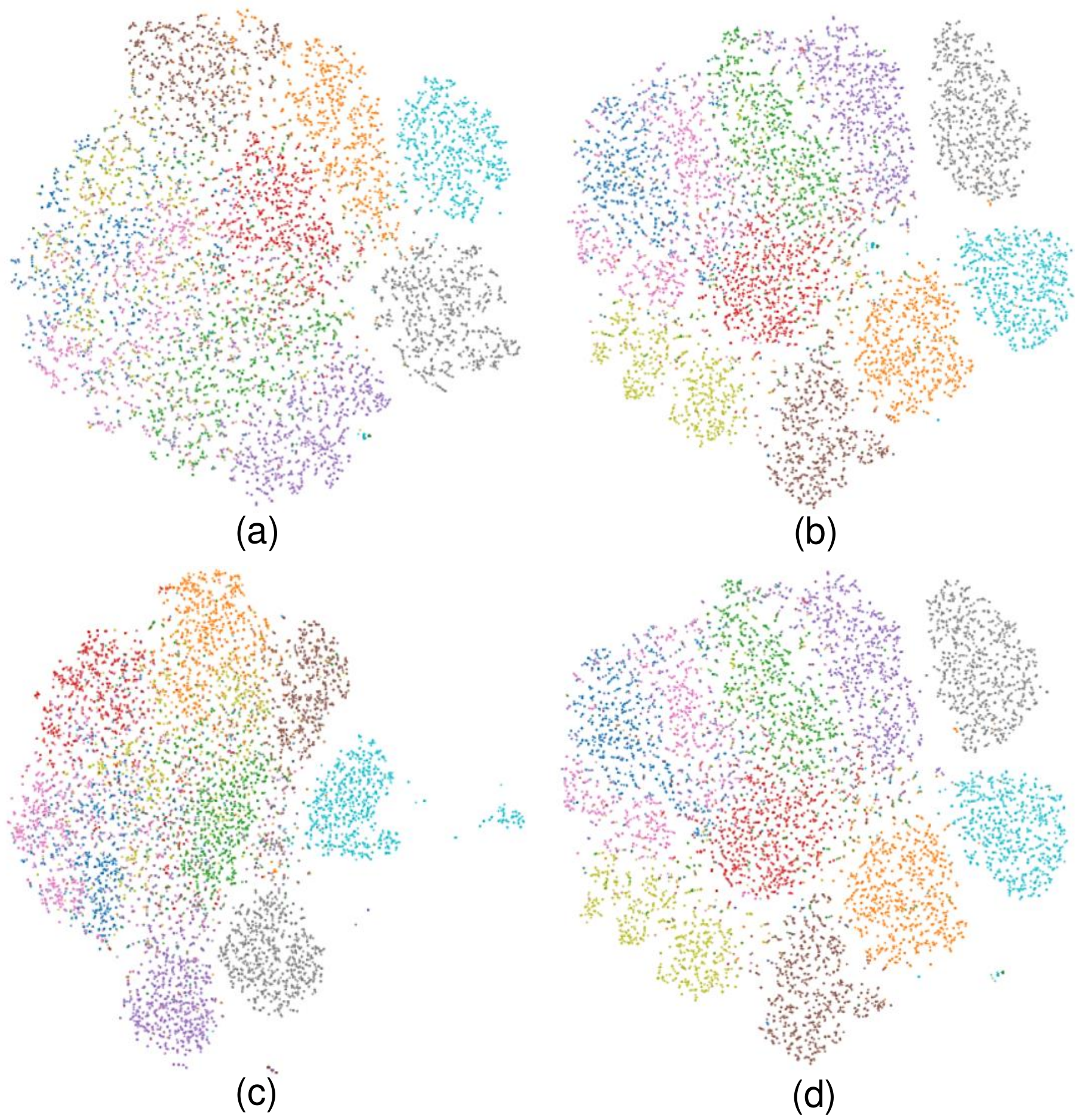}
\end{center}
\caption{TSNE visualizations for (a) Skeleton-DML under RA, (b) \textsc{Trans4SOAR}-Base under RA, (c) Skeleton-DML under RE and (d) \textsc{Trans4SOAR}-Base under RE on NTU-60~\cite{shahroudy2016ntu}.}
\label{fig:tsne_case}
\end{figure}
\subsection{Analyses Regarding Random Occlusion (RA)}
Random occlusion, considered as a combination between random temporal and spatial occlusions, is leveraged as the second main occlusion in our work on NTU-120~\cite{liu2019ntu}, NTU-60~\cite{shahroudy2016ntu} and Toyota Smart Home~\cite{Das_2019_ICCV}, depicted in Table~\ref{tab:ntu_120_occ} (b), Table~\ref{tab:ntu_60_all} (c) and Table~\ref{tab:tyt_all} (c), with $SNR=0.1$ and without occlusion on reference set respectively.
The proposed \textsc{Trans4SOAR} (Base) keeps surpassing all the existing approaches by large margins, \textit{e.g.}, SL-DML~\cite{memmesheimer2021sl} and Skeleton-DML~\cite{memmesheimer2020skeleton_dml} by $10.64\%$ and $18.02\%$ on NTU-120~\cite{liu2019ntu}.
The performance of Skeleton-DML~\cite{memmesheimer2020skeleton_dml} under RA is worse than that of SL-DML~\cite{memmesheimer2021sl}, while the case is reversed on RE, which means most of the existing approaches can not be robust against different occlusions.
However, \textsc{Trans4SOAR} overcomes this issue and demonstrates promising performances over different occlusions, especially for RE, which is an important ability for learning discriminative representation.
The proposed MAFM is also proved to have strong capability while dealing with different occlusions, which is well addressed through taking the three stream of skeleton patch embedding as input.
MAFM is further illustrated as the best fusion architecture among all the investigated fusion methods regarding the two main occlusions in Table~\ref{tab:ntu_60_fusion_comparison} which will be introduced later.
Considering the three streams encoding, first, since bone and velocity use temporal and spatial difference respectively, more cues regarding the neighbourhood of the occluded region could be encoded in different perspectives.
Furthermore, while tackling with RA on the NTU-60~\cite{shahroudy2016ntu} and Toyota Smart Home~\cite{Das_2019_ICCV}, \textsc{Trans4SOAR} (Base) and \textsc{Trans4SOAR} (Small) also demonstrate promising state-of-the-art performance across all the leveraged metrics, which reflects the strengths of the proposed models in multiple point of views. Similar ablations regarding the SNR ratio, \textit{i.e.}, $0.1$, $0.2$ and $0.3$, of the RA, are done in Table~\ref{tab:ntu_60_different_random_ratio}, where the performances of \textsc{Trans4DARC} (Base) and \textsc{Trans4DARC} (Small) surpass all the leveraged approaches. The experiments are done with occlusion on reference set. Especially for $SNR=0.1$ and $SNR=0.3$, \textsc{Trans4DARC} (Base) achieves $72.59\%$ and $72.39\%$ for accuracy while the Skeleton-DML~\cite{memmesheimer2020skeleton_dml} only achieves $60.43\%$ and $58.93\%$. Due to the large disruption by using $SNR=0.5$ with RA, the performance of \textsc{Trans4SOAR} (Base) only achieves $54.82\%$ accuracy while still outperforming the state-of-the-art approach by $10.13\%$.
\subsection{Analyses for Occlusion Disruption on Reference Samples}
Experiments are conducted in Table~\ref{tab:ntu_60_reference_ablation} to investigate different occlusions on the NTU-60~\cite{shahroudy2016ntu} reference set with the occlusion state denotes by OCCVal, where T and F indicate with occlusion and without occlusion. We set $SNR=0.1$ for RA and SNR range $0.05\rightarrow0.2$ for RE, which is comparable regarding averaged SNR. SL-DML~\cite{memmesheimer2021sl} and Skeleton-DML~\cite{memmesheimer2020skeleton_dml} have absolute performance fluctuation for accuracy by $3.46\%$ and $11.13\%$ for RA, and $2.61\%$ and $1.63\%$ for RE. What we desire is that the model should have small absolute fluctuation regarding different OCCVal setting. The experimental results of \textsc{Trans4SOAR} suit this desire with absolute fluctuation $1.00\%$ for RA and $0.80\%$ for RE, illustrating the strong ability against the occlusion on the reference set. Considering RA and RE with $OCCVal=F$, both the performances of SL-DML~\cite{memmesheimer2021sl} and Skeleton-DML~\cite{memmesheimer2020skeleton_dml} are worse with RE than that with RA, indicating that RE is more challenging.
\subsection{Analyses for Ablation of Fusion Mechanisms}
To demonstrate the efficiency of MAFM, we conduct comparison experiments among several fusion approaches in Table~\ref{tab:ntu_60_fusion_comparison}, where we set $SNR=0.1$ for RA and SNR range $0.05\rightarrow0.2$ for RE.
The mostly leveraged fusion technique is late fusion which conducts fusion at the decision level.
However the design of late fusion triplicates the model size as $113M$ while the others are at $40M$ level .
Here, we consider to propose a efficient fusion mechanism at patch embedding level which takes both the model performance and size into consideration.
The baselines for patch embedding level fusion includes the addition, multiplication and concatenation, which are directly executed after the acquisition of the patch embeddings for the three streams.
Another method we compared with is late fusion, which conducts addition after obtaining the final embeddings of the three streams and has a three times larger model size. 
The experimental results indicate that MAFM has great performance compared with all the leveraged patch-embedding level fusiom baselines and the competitive late fusion on the NTU-60~\cite{shahroudy2016ntu} under different occlusions, \textit{e.g.}, No occlusion (N), REalistic synthesized occlusion (RE) and RAndom occlusion (RA). Specifically, \textsc{Trans4SOAR} with MAFM surpasses the late fusion by $2.35\%$, $1.43\%$, and $17.04\%$ on the RE, RA, and N respectively, while having a smaller model size for both inference and training.
Simultaneously. \textsc{Trans4SOAR} with MAFM surpasses the investigated approach with the best performance among the basic patch embedding level fusion approach by $0.67\%$, $4.54\%$ and $6.44\%$ for RE, RA and N.

\subsection{Analyses for Random Temporal and Spatial Occlusions}
In order to show the performance of all the leveraged models with the two existing occlusions in the related work, \textit{i.e.}, random \emph{temporal} and \emph{spatial} occlusion, which might also be interesting to the community regarding the specific occlusion considering temporal and spatial components individually, we conducted experiments on three datasets while using the most effective approaches investigated in our work, as described in Table~\ref{tab:ntu_60_ramdomly_temporal_and_spatial_occ} (a) and Table~\ref{tab:ntu_60_ramdomly_temporal_and_spatial_occ} (b), where we choose the occluded frame number as $10$ and the occluded joints number as $5$ respectively. Compared with RE, these two leveraged occlusions which is specific controlled through predefined occluded frame and joint numbers are easier to be addressed as their randomness is not satisfied. But the important thing is that our proposed \textsc{Trans4SOAR} (Base) and \textsc{Trans4SOAR} (Small) still surpass all existing works by large margins on all datasets with these two occlusions which further illustrates the efficiency of our model against different occlusions, even the occlusion is specifically controlled by predefined concepts, \textit{e.g.}, the occluded frame number.

\begin{table}[t]
\centering
\caption{The comparison in terms of accuracy, the number of parameters (\#Params), and GFLOPs on NTU-120 without occlusion~\cite{liu2019ntu}}.
\label{tab:ntu_120_n_params}
\scalebox{0.8}{\begin{tabular}{lccc} 
\toprule
\textbf{Encoder} & \multicolumn{1}{l}{\textbf{Accuracy}} & \multicolumn{1}{l}{\textbf{\#Params}} & \multicolumn{1}{l}{\textbf{GFLOPS}} \\ 
\midrule
\multicolumn{4}{l}{\textbf{Previously Published Approaches}} \\ 
\midrule
AN$^\dagger$~\cite{liu2017global} & 41.0 & - & -  \\
FC$^\dagger$~\cite{liu2017global} & 42.1 & - & -  \\
AP$^\dagger$~\cite{liu2017global} & 42.9 & - & -  \\
APSR~\cite{liu2017global} & 45.3 & - & -  \\
TCN-OneShot~\cite{sabater2021one} & 46.3 & 3.5M & 8.5  \\
SL-DML~\cite{memmesheimer2021sl} & 50.9 & 11.2M & 23.8 \\
Skeleton-DML~\cite{memmesheimer2020skeleton_dml} & 54.2 & 11.2M & 23.8 \\ 
\midrule
\multicolumn{4}{l}{\textbf{CNN-based Encoder Optimized by DML}} \\ 
\midrule
SL-DML (AlexNet~\cite{krizhevsky2014one}) & 40.33 & 57.1M & 9.2\\
SL-DML (SqueezeNet~\cite{iandola2016squeezenet}) & 42.55 & 0.7M & 9.7\\
SL-DML (ResNet18~\cite{he2015deep}) & 49.19 & 11.2M & 23.8 \\
\midrule
\multicolumn{4}{l}{\textbf{GCN-based Encoder Optimized with DML (Ours)}}\\
\midrule
SL-DML (CTR-GCN~\cite{chen2021channel}) & 43.92 & 1.6M & 9.2\\
SL-DML (STTR~\cite{plizzari2021skeleton}) & 39.56 & 7.0M & 37.4\\
\midrule
\multicolumn{4}{l}{\textbf{Transformer-based Encoder Optimized with DML (Ours)}} \\ 
\midrule
SL-DML (CaiT~\cite{touvron2021cait}) & 47.86 & 120.8M& 53.9\\
SL-DML (ViT~\cite{dosovitskiy2020vit}) & 48.45 & 53.6M & 27.1 \\
SL-DML (Twins~\cite{chu2021twins}) & 49.00 & 25.2M & 75.1 \\
SL-DML (ResT~\cite{zhang2021rest}) & 52.58 & 57.8M & 61.3 \\
SL-DML (Swin~\cite{liu2021swin}) & 53.13 & 87.3M & 29.3 \\
SL-DML (LeViT~\cite{graham2021levit}) & 53.19 & 38.9M & 30.4\\ 
\midrule
\multicolumn{4}{l}{\textbf{Our Proposed and Extended Approaches (Ours)}} \\ 
\midrule
SL-DML (LeViT) + LSC & 55.94 & 38.9M & 30.4 \\
Trans4SOAR (Small) & 56.27 & 23.1M & 34.1\\
Trans4SOAR (Base) & \textbf{57.05} & 43.8M & 47.9\\
\bottomrule
\end{tabular}}
\end{table}

\subsection{Analysis for Qualitative and TSNE Experimental Results}
\noindent\textbf{Qualitative analysis. }The qualitative results are given in Figure~\ref{fig:qualitative} for SOAR with RE on the NTU-60~\cite{shahroudy2016ntu}, where the occluded body joints are marked as red dots. \textsc{Trans4SOAR} has overall great performance while comparing with Skeleton-DML~\cite{memmesheimer2020skeleton_dml} with true prediction $3:2$ among the selected $4$ samples. The occlusion of the joints which is dominant to the action has a large influence on the model, \textit{e.g.}, arm and hand joints for \emph{TakeOffGlasses} action, where Skeleton-DML~\cite{memmesheimer2020skeleton_dml} gives a false prediction while \textsc{Trans4SOAR} pursues a true prediction. However, due to the high similarity between several actions, \textit{e.g.}, \emph{WipeFace} and \emph{DrinkingWater}, \textsc{Trans4SOAR} still has false prediction but the predicted \emph{DrinkingWater} action is more similar with \emph{WipeFace} compared with \emph{TearUpPaper} predicted by Skeleton-DML~\cite{memmesheimer2020skeleton_dml}, showing that there is still research space for the future research.
We further present failure cases in Figure~\ref{fig:failure_case} to investigate the cause of the false prediction of our model for SOAR under RE. Since SOAR is only able to harvest informative classification cues from the information of the given fixed number of the human body joints, occlusion, which is assigned to the most dominant joint region during a specific movement type, causes large information decrease during the feature extraction procedure resulting in false prediction for SOAR. Considering the first sample in Figure~\ref{fig:failure_case}, the hand region is occluded when the person is picking something up, however the hand and arm region is the dominant region for the action throw. The information decrease on the dominant region makes our model predict pointing at something, which is a false prediction. Considering the second sample, most of the leg region are occluded by the projected object while the leg region is the dominant region during falling. The missing information also causes a negative effect on our model for the SOAR task.

\noindent\textbf{TSNE analysis. } In Figure~\ref{fig:tsne_case}, a TSNE~\cite{van2008visualizing} comparison among (a) Skeleton-DML under RA, (b) \textsc{Trans4SOAR}-Base under RA, (c) Skeleton-DML under RE, and (d) \textsc{Trans4SOAR}-Base under RE is shown to deliver a better understanding regarding the learned features on the latent space. First, compared with (a) and (c), (b) and (d) harvest clearer boundaries for more classes in the SOAR task, which showcases that \textsc{Trans4SOAR} has the capability to obtain embeddings with more discriminative cues. Then, if we look at the same approach under different occlusions, smaller changes are shown for \textsc{Trans4SOAR}-Base, as demonstrated in (b) and (d) while the the shape and structure of the latent space embeddings extracted from Skeleton-DML has larger changes as shown in (a) and (c). Overall, our \textsc{Trans4SOAR} approach shows better robustness against the different occlusions from the perspective of the change of the learned features on the latent space.

\subsection{Analyses for the Model Efficiency.}
To have a detailed look at the efficiency of different approaches, the accuracy for SOAR on no-occluded NTU-120, the number of the parameters, and the GFLOPS during inference are listed in Table~\ref{tab:ntu_120_n_params}. The number of parameters and GFLOPS for the first four approaches under \textit{Previously Published Approaches} are not available.
Compared with the visual transformer-based approaches, CNN-based approaches and GCN-based approaches preserve a smaller number of parameters and the GFLOPS while mostly delivering an unsatisfied performance for the SAOR task. 
The high performance of the visual transformer-based approaches are not exactly due to using larger models, since SL-DML (CaiT) has the largest number of the parameter and SL-DML (ResT) has the largest GFLOPS, but they do not achieve better performances compared with SL-DML (LeViT), which has $53.19\%$ in accuracy, $38.9M$ parameters, and $30.4$ GFLOPS. \textsc{Trans4SOAR} has a competitive amount of parameters and GFLOPS compared with other visual transformer approaches while delivering the best performance for the SOAR task. Especially, \textsc{Trans4SOAR} (Small) shows $56.27\%$ in accuracy with only $23.1M$ parameters and $34.1$ GFLOPS. Since our model is a multi-modality model, a reasonable increment in terms of the number of parameters and the GFLOPS is expected. \textsc{Trans4SOAR} (Base) achieves a $>70M$ parameter decrease compared with the late fusion approach while harvesting a better performance for SOAR, which demonstrates the superiority of \textsc{Trans4SOAR} from the perspective of multi-modality fusion.

\section{Conclusion}
In this work, we look into the problem of data-scarce recognition of daily activities through the lens of one-shot recognition, while considering diverse occlusions.
First, we propose realistic synthesized and random occlusion to better address the occlusion problem.
Then, a novel architecture \textsc{Trans4SOAR} is put forward to provide discriminative representations for skeleton input and enhance the robustness against different scenarios.
We design a \emph{Mixed Attention Fusion Mechanism (MAFM)}, featuring a three-stream of skeleton encoding inputs to realize efficient fusion on the patch-embedding level.
Inspired by recent success of augmentation-based methods in semi-supervised learning, we further introduce the latent space consistency loss, which leverages an additional auxiliary branch encouraging the embedder to produce similar results despite extensive augmentations at the feature level.
\textsc{Trans4SOAR} sets the new state of the art on both normal and occluded SOAR benchmarks established on three datasets. In the future, occluded one-shot recognition based on video data is still attractive to be researched.

\bibliographystyle{IEEEtran}
\bibliography{bib}

\end{document}